\documentclass{article}
\usepackage[final]{neurips_2023}
\usepackage{bbding}
\usepackage{float}
\usepackage{natbib}
\setcitestyle{numbers,square}

\usepackage[utf8]{inputenc} 
\usepackage[T1]{fontenc}    
\usepackage{hyperref}       
\usepackage{url}            
\usepackage{booktabs}       
\usepackage{amsfonts}       
\usepackage{nicefrac}       
\usepackage{microtype}      
\usepackage{multirow}
\usepackage{CJKutf8}
\usepackage[many]{tcolorbox}
\usepackage{graphicx}
\usepackage{geometry}
\usepackage{tabularx}

\usepackage{makecell}
\usepackage{cleveref}
\usepackage[table]{xcolor}

\usepackage{threeparttable}
\usepackage{subcaption} 

\usepackage{wrapfig}

\hypersetup{
    colorlinks=true,
    linkcolor=blue,
    citecolor=blue,
    urlcolor=blue,
    pdftitle={MonkeyOCRv2: A Visual-Text Foundation Model for Document AI},
}

\definecolor{avgcolor}{RGB}{235,245,255}      
\definecolor{overallcolor}{RGB}{255,240,220}  

\renewenvironment{abstract}{%
  \vskip 0.075in%
  \centerline{\large\bf Abstract}%
  \vskip 0.075in%
  \noindent\ignorespaces%
}{%
  \vskip 0.1in%
}

\definecolor{green}{RGB}{0,150,10}
\definecolor{blue}{RGB}{0,148,181}
\definecolor{orange}{RGB}{194,153,107}

\usepackage{fancyhdr}
\usepackage{fancyhdr}

\definecolor{yl_color}{RGB}{128, 255, 0}

\fancypagestyle{firstpage}{%
  \fancyhf{}
}

\fancypagestyle{otherpage}{%
  \fancyhf{}

}

\title{MonkeyOCRv2: A Visual-Text Foundation Model for Document AI}

\author{
Yuliang Liu\textsuperscript{1\thanks{Equal contribution.}},
Zhang Li\textsuperscript{1\footnotemark[1]},
Ziyang Zhang\textsuperscript{1\footnotemark[1]},
\textbf{Shuo Zhang\textsuperscript{1\footnotemark[1]}},
Qiang Liu\textsuperscript{2},
\textbf{Jiajun Song\textsuperscript{1}},\\ 
 \textbf{Zidun Guo\textsuperscript{1}}, 
 \textbf{Xinhan Wang\textsuperscript{1}}, \textbf{Handong Zheng\textsuperscript{1}},
 \textbf{Yang Liu\textsuperscript{1}},  \textbf{Dongliang Luo\textsuperscript{1}}, 
 \\ \textbf{Zhiyin Ma\textsuperscript{1}}, \textbf{Jiarui Zhang\textsuperscript{2}}, \textbf{Xiang Bai\textsuperscript{1}} \\ 
[0.4em] 
\textsuperscript{1}Huazhong University of Science and Technology, \textsuperscript{2}Kingsoft Office\\[0.4em] 
  {\normalfont\ttfamily\small\begin{tabular}[t]{@{}c@{}}
  \{ylliu, zhangli123, zzyzz, xbai\}@hust.edu.cn
  \end{tabular}}
}

\begin{document}

\maketitle
\thispagestyle{firstpage}  

\begin{abstract}
    Mainstream visual encoders are pretrained on natural images and cannot be effectively applied to document images without document-oriented adaptation, as dense text and fine-grained character strokes demand character-level visual perception. We present MonkeyOCRv2, a visual-text pretrained model for document AI. First, we construct MonkeyDoc v2, to our knowledge the largest document-image pretraining corpus, comprising 113 million images spanning 17 languages. 
    Second, we propose a pretraining strategy that jointly learns image-to-text generation and pixel-level document reconstruction: the former aligns visual representations with textual content, while the latter preserves character strokes and layout details. Extensive experiments are conducted on five representative document analysis tasks, including text recognition, formula recognition, text detection, document tampering detection, and overlapping text segmentation. Replacing the original encoders with MonkeyOCRv2 consistently improves performance across all five tasks, raising the overall recognition accuracy of CRNN from 58.7\% to 67.3\% and enabling the 110M UniMERNet-T to outperform the 325M UniMERNet-B.
    Finally, we validate its effectiveness as the vision encoder of multimodal large language models on the more challenging tasks of document parsing and document understanding. Kept frozen and paired with a lightweight language model, it yields a 0.7B document parsing model that sets a new open-source state-of-the-art on MDPBench, a recent benchmark spanning digital-born and photographed documents across 17 languages, surpassing the previous best 3B dots.mocr by 2.8\% absolute with a vision encoder roughly 11$\times$ smaller; on OmniDocBench, it further outperforms much larger general-purpose VLMs such as Qwen3-VL-235B and GPT-5.2. The frozen encoder also powers a document understanding model that outperforms counterparts built on CLIP, DINO, and SAM across eight benchmarks under identical training settings. 
    These results suggest that document-oriented visual pretraining can serve as a foundation for document intelligence in its own right. Code and data will be released at \url{https://github.com/Yuliang-Liu/MonkeyOCRv2}.
\end{abstract}

   \begin{figure*}[t]
    \centering
    \includegraphics[width=1\textwidth]{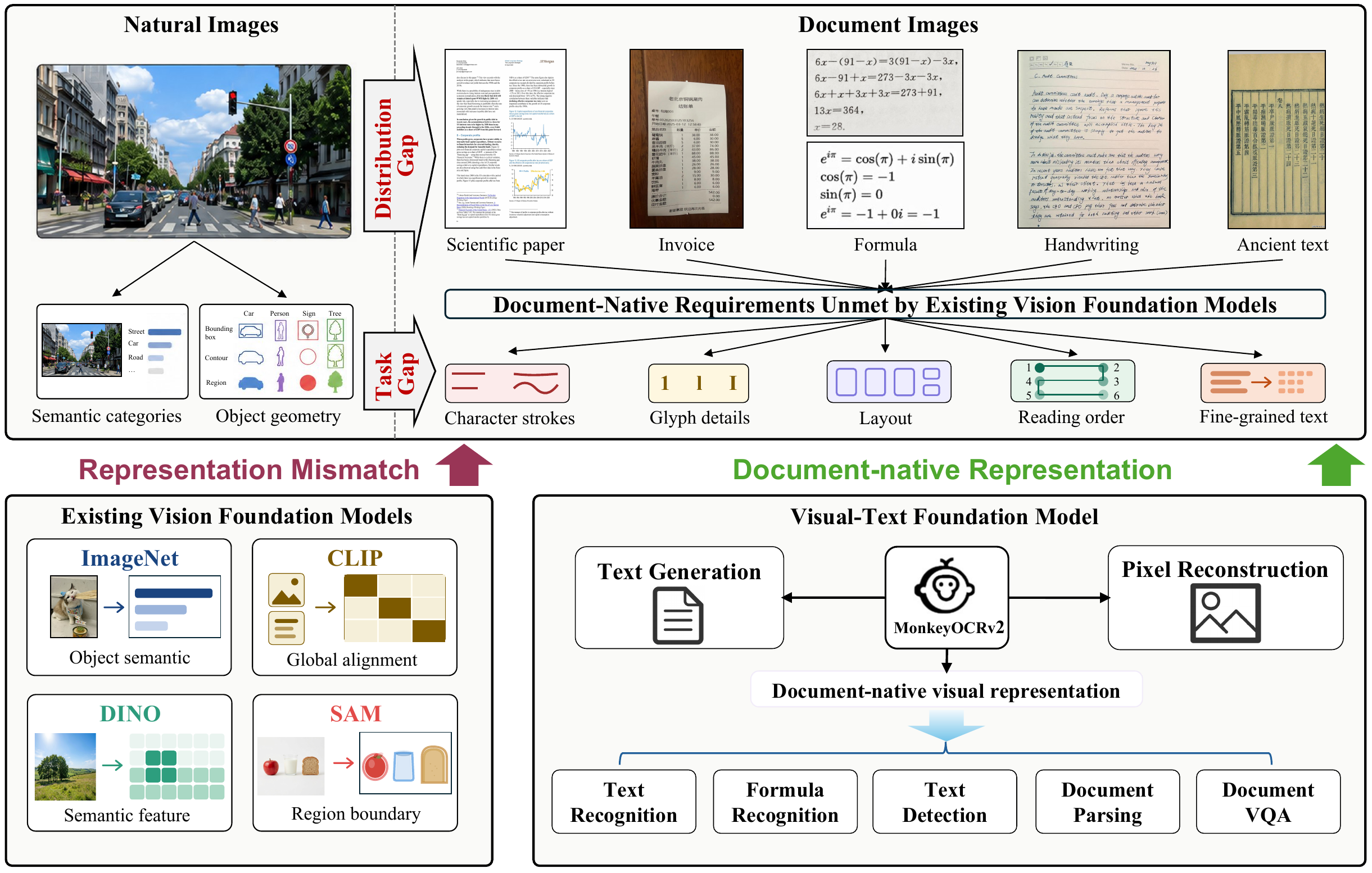}
        \caption{\textbf{Overview of MonkeyOCRv2.} Existing vision foundation models are primarily designed for natural images and emphasize object semantics, global alignment, semantic features, or region boundaries. MonkeyOCRv2 addresses the resulting representation mismatch by jointly learning text generation and pixel-level reconstruction, producing document-native visual representations that transfer across diverse document AI tasks.}
        \label{fig:overview}
\end{figure*}

\setcounter{footnote}{0}

\section{Introduction}
Human knowledge is extensively preserved in visual-text format across various document images, including scientific papers, technical standards, legal instruments, financial reports, medical records, handwritten notes, and corporate files. Accurately perceiving, parsing, and understanding the textual and visual information within these documents is fundamental to knowledge extraction, document digitization, intelligent enterprise office systems, corpus construction for large language model training, and AI for Science~\cite{ai4s_1, ai4s_2}. In recent years, large-scale pre-training has driven rapid advances in computer vision. General representations learned by pretrained visual encoders can be effectively transferred to diverse downstream tasks, yielding substantial performance gains. Nevertheless, popular foundational vision models such as CLIP~\cite{clip}, SAM~\cite{sam,sam2} and DINO~\cite{dino,dinov2,dinov3} are primarily trained on natural images. Their data distributions and pre-training objectives differ drastically from document scenarios, rendering them ill-suited for document parsing, document understanding, and related tasks.

Document images are characterized by dense text, fine-grained character structures, and complex layouts, with their semantics heavily dependent on local visual details. Even subtle variations in character strokes, local structures, punctuation marks, decimal points, or superscripts/subscripts can lead to entirely different meanings.
In contrast, natural-image understanding primarily focuses on high-level semantics, including object categories, scene context, and visual concepts, and is generally invariant to semantically irrelevant local appearance changes.

\begin{figure*}[t]
    \centering
    \begin{minipage}[t]{0.58\textwidth}
        \centering
        {\scriptsize\textbf{(a) Multilingual document parsing on MDPBench}\par}
        \vspace{0.2em}
        \includegraphics[width=\linewidth]{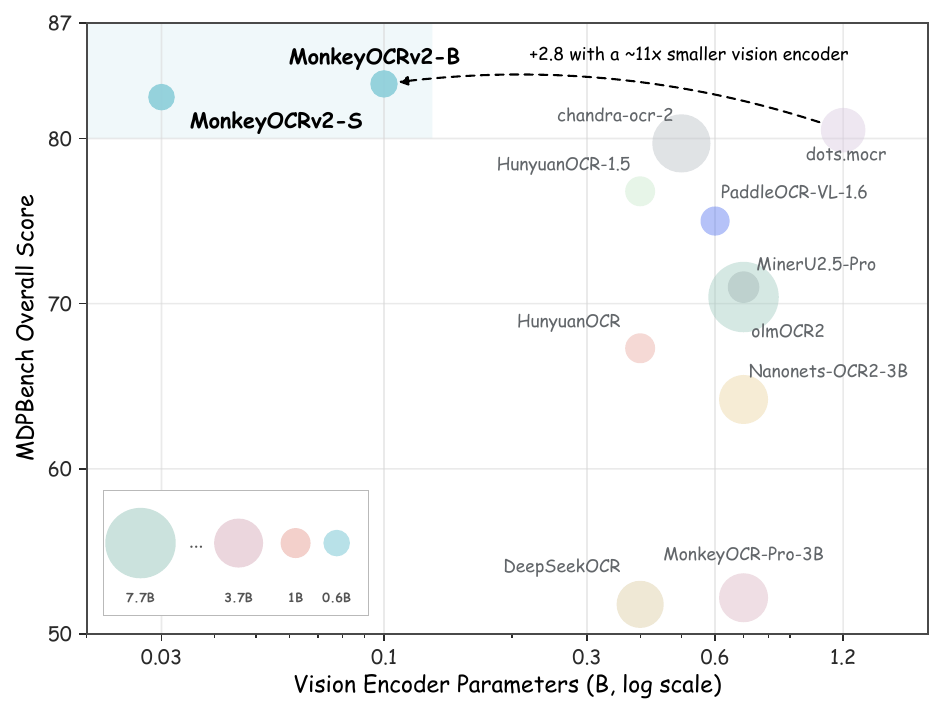}
    \end{minipage}\hfill
    \begin{minipage}[t]{0.39\textwidth}
        \centering
        {\scriptsize\textbf{(b) Consistent gains across 7 document analysis tasks}\par}
        \vspace{0.2em}
        \includegraphics[width=\linewidth]{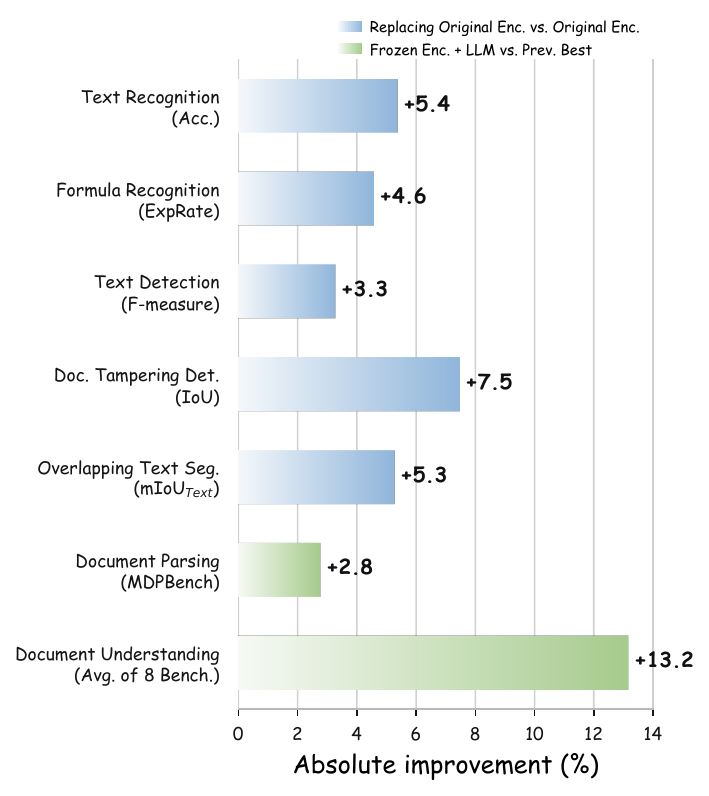}
    \end{minipage}
        \caption{\textbf{Performance overview of MonkeyOCRv2.} (a) Performance versus vision-encoder size on MDPBench~\cite{mdpbench}, a challenging multilingual document parsing benchmark. MonkeyOCRv2 achieves 83.3\%, outperforming the previous best open-source model, dots.mocr, with a vision encoder roughly 11$\times$ smaller. Bubble area indicates the total number of model parameters. (b) Absolute performance improvements across seven document  analysis tasks. Blue bars show the gains obtained by replacing the original encoders with MonkeyOCRv2 and fine-tuning the downstream models; results are averaged across downstream architectures when multiple architectures are evaluated. Green bars show the gains obtained by keeping MonkeyOCRv2 frozen and pairing it with a lightweight LLM. For document understanding, the improvement is measured against the best-performing baseline encoder under identical training settings.
        }
        \label{fig:performance_overview}
\end{figure*}

This discrepancy is further reflected in pre-training objectives. Most popular vision pretrained models are designed around category-level, object-level, or scene-level semantic modeling. ImageNet~\cite{imagenet} classification pre-training aims to learn category-discriminative representations, encouraging the model to disregard local variations in pose, texture, and background; vision-language models such as CLIP~\cite{clip} and SigLIP~\cite{siglip} emphasize global semantic alignment between images and text; self-supervised methods like DINO~\cite{dino} learn stable semantic representations through cross-view consistency; and SAM~\cite{sam} focuses primarily on modeling region boundaries and segmentable objects. While these methods deliver outstanding performance on natural image tasks, they all share the trait of prioritizing high-level semantic information over preservation of character-level fine-grained visual differences.

To fill this gap, we present MonkeyOCRv2, a visual foundation model for document AI: an encoder pretrained on document images with objectives that reward character-level visual fidelity rather than global semantic abstraction. As illustrated in Fig.~\ref{fig:overview}, MonkeyOCRv2 addresses the representation mismatch between natural-image encoders and document images through joint text generation and pixel-level reconstruction, and transfers to diverse document AI tasks.

First, we construct MonkeyDoc v2, to our knowledge the largest document-oriented visual-text pretraining dataset, comprising 113 million images across 17 languages (Tab.~\ref{tab:foundation_data_comparison}). Existing visual encoders are pretrained primarily on natural-image datasets such as ImageNet~\cite{imagenet} and SA-1B~\cite{sam}, which offer at best coarse textual annotations and thus lack the dense supervision that document understanding requires. Moreover, the training data of widely used models such as CLIP, SigLIP, and DINOv2 are not publicly available, making it difficult to adapt or extend their pretraining for document-centric applications. Existing document-oriented pretraining datasets are similarly limited: the training data of oCLIP~\cite{oclip} consist mainly of scene-text images; DiG~\cite{dig} is dominated by small-scale scene text recognition data; and IIT-CDIP~\cite{iitcdip}, the corpus behind DiT~\cite{dit}, lacks fine-grained text annotations and diversity in document types. Furthermore, these datasets generally cover only Chinese and English, providing limited multilingual support.

Second, we propose a pretraining strategy for document images that couples image-to-text generation with pixel-level document reconstruction. The generation objective aligns visual representations with their corresponding textual content, providing direct and dense supervision. In parallel, the reconstruction objective encourages the encoder to preserve fine-grained visual details, including character strokes, glyph shapes, and layout structures, that may be discarded under purely textual supervision. This complementary supervision strengthens the grounding of downstream predictions in visible evidence, particularly when linguistic context is weak or unavailable. A controlled scrambled-text study (Sec.~\ref{subsec:alpd}) supports this interpretation: at low resolution, reconstruction nearly halves the accuracy gap between semantically coherent and scrambled text. We use this gap as an operational proxy for dependence on linguistic context.

As summarized in Fig.~\ref{fig:performance_overview} (b), we evaluate MonkeyOCRv2 on five representative document  analysis tasks: text recognition, formula recognition, text detection, document tampering detection, and overlapping text segmentation. Replacing the original visual encoders with MonkeyOCRv2 yields consistent gains across all five tasks: it improves text recognition by an average of 5.4\% absolute on challenging English, Chinese, and occluded-text benchmarks, enables a 110M formula recognition model to surpass the 325M counterpart, and brings improvements of 3.3\%, 7.5\%, and 5.3\% on text detection, document tampering detection, and overlapping text segmentation, respectively.

\begin{table}[t]
\centering
\caption{Comparison of the training corpora of representative visual pretrained models. Existing encoders are pretrained on natural images or narrowly scoped text data. To the best of our knowledge, MonkeyDoc v2 is the largest corpus for document-oriented visual pretraining, covering a broad spectrum of document types, including printed papers, scanned books, handwritten notes, newspapers, magazines, financial reports, rendered documents, and more.}
\label{tab:foundation_data_comparison}
\begin{tabular}{@{}lrlc@{}}
\toprule
\textbf{Model} & \textbf{Data Scale} & \textbf{Data Type} & \textbf{Languages} \\
\midrule

\multicolumn{4}{l}{\textit{General visual pretrained models}} \\
\midrule
CLIP~\cite{clip}        & 400M   & Natural images & N/A \\
SigLIP~\cite{siglip}    & 10B    & Natural images & N/A \\
SigLIP 2~\cite{siglip2} & $>$40B & Natural images & N/A \\
SAM~\cite{sam}          & 11M    & Natural images & N/A \\
SAM 2~\cite{sam2}       & 50.9K  & Natural videos & N/A \\
DINO~\cite{dino}        & 1.28M  & Natural images & N/A \\
DINOv2~\cite{dinov2}    & 142M   & Natural images & N/A \\
DINOv3~\cite{dinov3}    & 1.7B   & Natural images & N/A \\

\midrule
\multicolumn{4}{l}{\textit{Document-oriented visual pretrained models}} \\
\midrule
oCLIP~\cite{oclip}      & 450K   & Scene-text images & 2 \\
DiG~\cite{dig}          & 35.6M  & Scene-text images & 1 \\
DiT~\cite{dit}          & 42M    & Scanned documents & 1 \\
Donut~\cite{donut}      & 13M & Scanned / synthetic documents & 4 \\
Pix2Struct~\cite{pix2struct} & 80M & Webpage screenshots & 1 \\
\addlinespace[2pt]
\textbf{MonkeyOCRv2}    & \textbf{113M} & \textbf{Multi-type documents} & \textbf{17} \\

\bottomrule
\end{tabular}
\end{table}

Finally, we investigate MonkeyOCRv2 on two more challenging tasks: document parsing and document understanding. Following the prevailing paradigm, we combine the frozen encoder with large language models to build a 0.7B document parsing model, MonkeyOCRv2-Parsing, and a document understanding model. As shown in Fig.~\ref{fig:performance_overview} (a), despite its lightweight 0.1B visual encoder, MonkeyOCRv2-Parsing achieves state-of-the-art performance among open-source models on MDPBench~\cite{mdpbench}, a challenging multilingual document parsing benchmark. It surpasses PaddleOCR-VL-1.6~\cite{ppocrvl16} by 8.3\% and the previous best open-source model, the 3B dots.mocr~\cite{dotsmocr}, by 2.8\%, while using a vision encoder roughly 5$\times$ and 11$\times$ smaller, respectively.
For document understanding, we adopt the conventional VLM framework~\cite{llava1.5} and compare pretrained visual encoders under identical training data and settings, with all encoders kept frozen; our model consistently outperforms counterparts built on CLIP, DINO, SAM, and previous document-oriented pretrained encoders across eight widely used benchmarks.

Our contributions are as follows:
\begin{itemize}
    \item We propose MonkeyOCRv2, a visual-text pretrained encoder for document AI. 
    Its two-pronged pretraining couples image-to-text generation with pixel-level document reconstruction: the former aligns visual representations with textual content, while the latter retains fine-grained visual evidence and improves robustness when linguistic context is weak or unavailable.
    \item We construct MonkeyDoc v2, to our knowledge the largest visual-text pretraining dataset for document scenarios, comprising 113 million densely annotated images across 17 languages.
    \item Extensive experiments validate the strong transferability of MonkeyOCRv2. As a backbone substitution, it brings consistent improvements across five representative document analysis tasks; kept frozen and paired with a 0.6B LLM, it achieves state-of-the-art open-source performance on MDPBench, a challenging multilingual document parsing benchmark, and outperforms mainstream pretrained encoders across eight document understanding benchmarks. 
    A controlled scrambled-text study further shows that the reconstruction objective improves recognition when linguistic context is removed and narrows the accuracy gap between semantically coherent and scrambled text.
\end{itemize}

\section{Related Work}
\label{sec:rela}

\subsection{General Visual Pretrained Models}
    Large-scale visual pretraining has become a central paradigm in computer vision, enabling encoders to learn transferable representations from massive datasets~\cite{pre9, pre6}. Early vision models were commonly pretrained in a supervised manner on annotated datasets such as ImageNet~\cite{imagenet}, which encouraged category-level visual discrimination and provided strong initialization for downstream tasks. More recently, self-supervised and vision-language pretraining have significantly advanced vision foundation models. Vision-language models such as CLIP~\cite{clip} and SigLIP~\cite{siglip,siglip2} align images with natural language descriptions at scale, leading to strong global semantic understanding and zero-shot transferability. Self-supervised methods such as DINO~\cite{dino,dinov3} learn general-purpose features through self-distillation and cross-view consistency, showing strong performance in dense prediction and representation transfer. In parallel, SAM~\cite{sam} introduces a promptable segmentation framework trained on the large-scale SA-1B dataset, demonstrating impressive generalization to generic object and region segmentation.

    Recent studies further explore unified and scalable pretraining paradigms. OpenVision~\cite{openvision} combines contrastive learning with generative supervision to improve semantic modeling, while OpenVision 2~\cite{openvision2} adopts a purely generative image-to-text paradigm by removing the text encoder and contrastive objectives. The RADIO series~\cite{radio,radio2d5} leverages multi-teacher distillation to integrate knowledge from diverse vision models into a single encoder. Despite their strong generalization, these general-purpose foundation models are designed for natural-image understanding, global semantic alignment, or generic region perception; none is optimized for the character-level visual discrimination that document images demand.

\subsection{Document-oriented Visual Pretrained Models}
    Early optical character recognition (OCR) models for document AI~\cite{dbnet} typically adopt ImageNet-pretrained vision encoders followed by task-specific optimization. To improve visual representation learning for OCR, several pretraining paradigms have been explored. DiG~\cite{dig} combines contrastive learning with masked image modeling for self-supervised pretraining, improving text recognition, text segmentation, and text image super-resolution. oCLIP~\cite{oclip} introduces a character-aware vision-language pretraining framework that leverages weakly annotated text to learn scene text representations, leading to significant gains in text detection. DiT~\cite{dit} exploits large-scale unlabeled document images for self-supervised pretraining and performs strongly on document image classification, layout analysis, table detection, and text detection. UniRec-0.1B~\cite{unirec} is a vision-language model trained on 40 million samples, supporting text and formula recognition across granularities from characters to full documents. Donut~\cite{donut} trains an OCR-free model consisting of a Swin Transformer~\cite{swin} encoder and a decoder on approximately 13 million document images to directly convert document images into structured text, while Nougat~\cite{nougat} builds upon Donut to process scientific documents into a markup language. Pix2Struct~\cite{pix2struct} trains an image-to-text model on masked web screenshots to parse visual language inputs into simplified HTML. TrOCR~\cite{trocr} uses a BEiT~\cite{beit} initialized transformer encoder to perform end-to-end optical character recognition.
    LayoutLMv3~\cite{layoutlmv3} is pretrained on 11M IIT-CDIP~\cite{iitcdip} documents to learn unified text, layout, and image representations for document understanding tasks.
    However, each of these efforts targets a narrow slice of document AI (recognition, detection, or layout) and their training data remain confined largely to English and Chinese with limited image diversity, falling short of a general-purpose, multilingual document encoder.

    With the rapid advancement of large language models and multimodal large language models, OCR research has evolved toward higher-level document AI tasks, including document parsing and document understanding~\cite{monkey,docowl1.5}. However, many existing systems still rely on general-purpose visual foundation models as their vision backbones: Qwen3-VL~\cite{qwen3vl} and HunyuanOCR series~\cite{hunyuanocr, hunyuanocr15} adopt SigLIP 2~\cite{siglip2}, TextMonkey~\cite{textmonkey} uses OpenCLIP's ViT-bigG~\cite{openclip}, DeepSeek-OCR~\cite{deepseekocr} is built upon SAM~\cite{sam}, and LLaVAR~\cite{llavar} utilizes CLIP~\cite{clip}.
    Vary~\cite{vary} uses the SAM-pretrained ViTDet~\cite{vitdet} image encoder and GOT-OCR2.0~\cite{gotocr2} follows Vary.

    More recent document parsing models instead initialize their vision encoders from multimodal large language models: the PaddleOCR-VL series~\cite{ppocrvl, ppocrvl15, ppocrvl16} adopts a 0.6B encoder initialized from Keye-VL~\cite{keyevl}, while the MinerU2.5 series~\cite{mineru25, mineru2.5pro} employs a 675M encoder initialized from Qwen2-VL~\cite{qwen2vl}. These backbones are pretrained on massive non-public data, making them difficult to reproduce and limiting their accessibility to the research community. 

    Our work differs from these efforts in three respects. First, unlike recent end-to-end document VLMs, we pretrain from scratch a standalone encoder intended as a backbone substitution and evaluate it as such across seven document  analysis tasks spanning recognition, detection, segmentation, tampering localization, parsing, and understanding.  
    Second, unlike generation-only pretraining, our objective adds a reconstruction term that encourages the encoder to preserve local visual evidence.
    Third, MonkeyDoc v2 extends document-oriented pretraining to 113M images across 17 languages.

\begin{figure*}[t]
    \centering
    \includegraphics[width=1\textwidth]{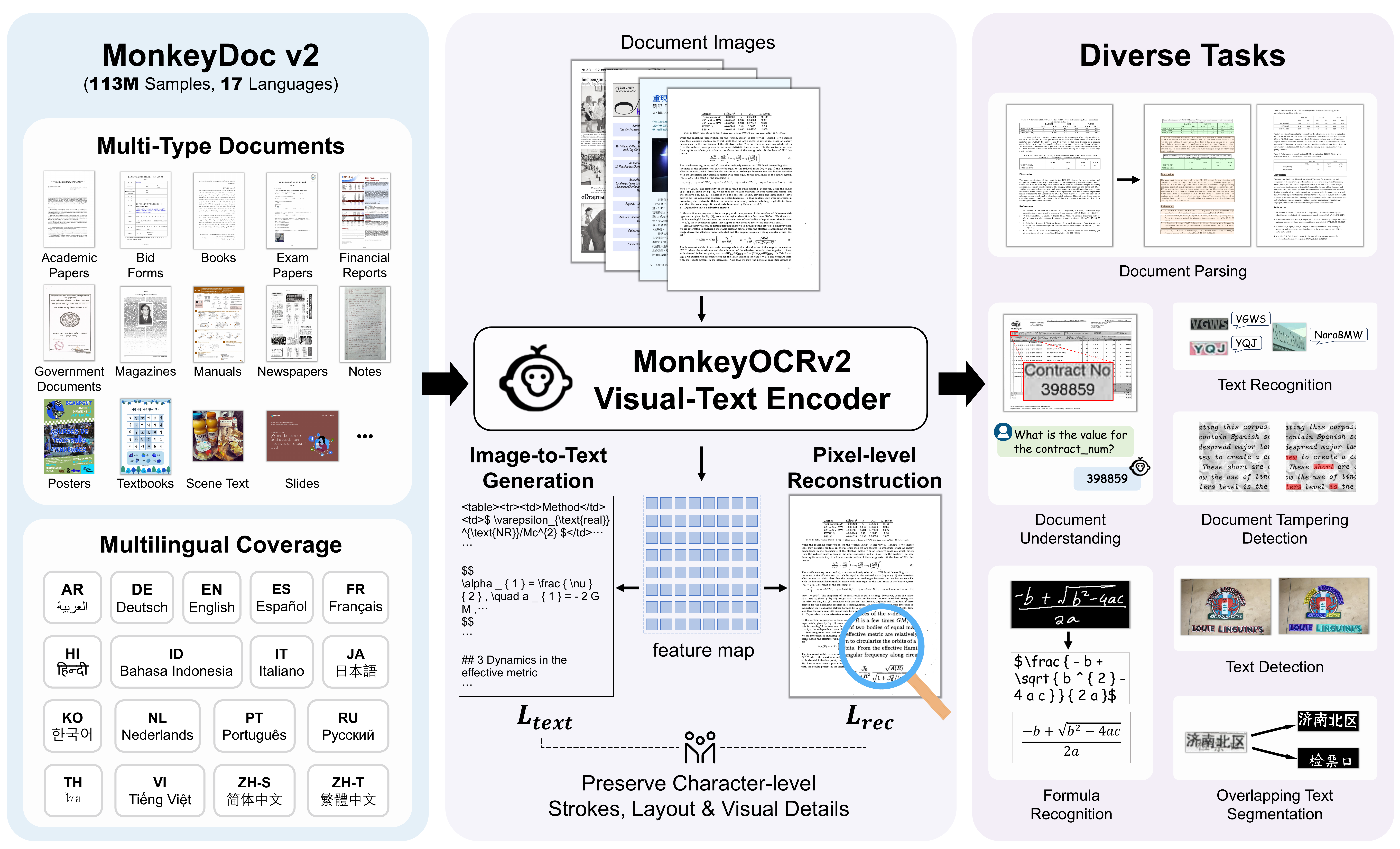}
        \caption{MonkeyOCRv2 is pretrained on a large-scale corpus of multilingual, multi-type document images, enabling strong generalization across diverse downstream document analysis tasks.}
        \label{fig:pre}
\end{figure*}

\section{MonkeyOCRv2}
\subsection{Data Engine}

To support document-oriented visual representation learning, we construct a large-scale multilingual pretraining dataset, termed \textbf{MonkeyDoc v2}. MonkeyDoc v2 contains 113 million samples and covers 17 languages\footnote{Following common practice in multilingual
NLP (e.g., FLORES-200), Simplified and Traditional Chinese are counted
as separate entries}, namely Simplified Chinese, Traditional Chinese, English, Arabic, German, Spanish, French, Hindi, Indonesian, Italian, Japanese, Korean, Dutch, Portuguese, Russian, Thai, and Vietnamese. 
The corpus contains 8M page-level images and 105M cropped document elements. Throughout this paper, a ``sample'' refers to either a page image or a cropped element paired with its supervision. Real and synthetic samples account for 61M samples (54\%) and 52M samples (46\%), respectively. 
App.~\ref{sec:appendix_data} reports detailed sample counts by source, language setting, data granularity, and task category.
The data engine consists of three core modules: Expert Model Labeling, Multilingual Corpus-Based Data Synthesis, and Data Filtering. In our framework, large language models and expert OCR systems are employed solely to provide more accurate annotations for real-world documents at scale.

\textbf{Expert Model Labeling:} Real-world documents are annotated through a multi-expert agreement pipeline rather than by any single model, and thus the supervision quality does not hinge on the idiosyncratic errors of one system. We first apply an off-the-shelf layout-detection model to perform document layout analysis and crop document elements, including text blocks, tables, formulas, and other regions. Each cropped element is then independently transcribed by several complementary expert recognition models with differing architectures and training data. For each element, we compute pairwise similarities among all expert predictions and retain the one with the highest average agreement, which suppresses model-specific failure modes and yields more reliable labels.
The complete toolchain (layout detector and expert recognizers) is enumerated in the released repository.

\textbf{Multilingual Corpus-Based Data Synthesis:} To enhance multilingual coverage, we synthesize large-scale OCR training data from multilingual corpora. Specifically, we randomly sample textual content from LLM corpora covering 17 languages and render it into document images with diverse fonts, styles, and resolutions. To better cover rare characters and low-frequency symbols, we extract the complete character set of each language and generate random character combinations for rendering.
For table data, we populate multilingual text into both real table templates and automatically generated table structures, enabling our model to learn diverse layouts, structural patterns, and spanning relationships.
For formula data, we collect formulas from papers crawled from arXiv and render them into image-text pairs, yielding approximately 0.8M formula samples.

\textbf{Data Filtering:} To further improve data quality, we filter low-quality annotated samples from challenging sources such as newspapers and handwritten notes. 
We first verify the completeness of layout annotations by masking all detected regions in a document image with white blocks and feeding the masked image into a strong document-oriented multimodal large language model. If the model can still recognize residual text, the sample is considered to contain missed detections or incomplete layout annotations and is therefore discarded. We then validate the logical consistency of the annotated reading order: for each sample, the recognized text is concatenated according to the annotated order and assessed by a large language model to determine whether reading-order errors or semantic inconsistencies remain. 
As above, these models serve only as annotation filters; the contribution lies in the filtering criteria rather than the specific judge models.
Through this filtering process, approximately 0.9 million of the initial 1.2 million pages from these challenging sources are retained.

\textbf{Data Source.} The raw data used to construct MonkeyDoc v2 are collected from a diverse set of document datasets, using only the official training splits of each dataset, including FinePDFs\footnote{\url{https://huggingface.co/datasets/HuggingFaceFW/finepdfs}}, MonkeyDoc~\cite{monkeyocr}, Union14M~\cite{union14m}, UniMER-1M~\cite{unimernet}, CDLA\footnote{\url{https://github.com/buptlihang/CDLA}}, D4LA~\cite{d4la}, DocGenome~\cite{docgenome}, DocLayNet~\cite{doclaynet}, M6Doc~\cite{m6doc}, SVRD~\cite{svrd}, TabRecSet~\cite{TabRecSet}, COCOTextV2~\cite{COCOTextV2}, HierText~\cite{HierText}, DSTextV2~\cite{DSTextV2}, LSVT~\cite{LSVT}, OpenImagesV5Text~\cite{OpenImagesV5Text}, ReCTS~\cite{ReCTS}, TextOCR~\cite{TextOCR}, and MTWI~\cite{MTWI}, as well as large-scale collections of newspapers, handwritten notes, and presentation slides gathered by us. 
For synthetic data generation, we leverage multilingual corpora from publicly available LLM training datasets~\cite{llmyuliao1,llmyuliao2}, Unicode character sets, 
and formulas crawled from arXiv papers; the resulting synthetic data account for 46\% of the overall corpus.
These resources enable the construction of large-scale multilingual OCR training data with diverse languages, layouts, structures, visual styles, and character distributions.
All official validation and test splits of downstream benchmarks are excluded from pretraining, downstream training, checkpoint selection, prompt selection, and hyperparameter tuning.

\subsection{Pre-training}
The pre-training framework consists of MonkeyOCRv2 vision encoder $E_v$, a vision decoder $D_v$, and a text decoder $D_t$. Given an input image $I \in \mathbb{R}^{H \times W \times 3}$, the vision encoder first extracts a sequence of visual tokens:
\begin{equation}
\mathbf{z} = E_v(I),
\end{equation}
where $\mathbf{z}$ denotes the visual tokens. The encoded visual tokens are then fed into the vision decoder to reconstruct the original image:
\begin{equation}
\hat{I} = D_v(\mathbf{z}).
\end{equation}

\paragraph{Reconstruction Objective.}
We reconstruct the input image from the visual tokens to encourage the encoder to preserve sufficient visual information for recovering fine-grained document content.
By default, we use a mean squared error (MSE) objective:
\begin{equation}
\mathcal{L}_{\mathrm{pix}}
=
\frac{1}{3HW}
\left\|
\hat{\mathbf{I}}-\mathbf{I}
\right\|_2^2,
\label{eq:pixel_loss}
\end{equation}
where $\mathbf{I}\in\mathbb{R}^{H\times W\times 3}$ and
$\hat{\mathbf{I}}$ denote the input and reconstructed images, respectively.

To further preserve stroke-level structures, we additionally investigate a
structure-aware reconstruction objective that matches the edge and
distance-to-edge representations of $\mathbf{I}$ and $\hat{\mathbf{I}}$.
For an RGB image $\mathbf{I}$, we first convert it to grayscale using
$f(\cdot)$ and compute its Sobel gradient magnitude:
\begin{equation}
\mathcal{G}(\mathbf{I})
=
\sqrt{
\left(
\mathbf{K}_{x} * f(\mathbf{I})
\right)^2
+
\left(
\mathbf{K}_{y} * f(\mathbf{I})
\right)^2
},
\label{eq:sobel_magnitude}
\end{equation}
where $\mathbf{K}_{x}$ and $\mathbf{K}_{y}$ are the horizontal and
vertical Sobel kernels, respectively, and $*$ denotes convolution.
The corresponding soft edge map is defined as
\begin{equation}
\mathcal{E}(\mathbf{I})
=
\sigma
\left(
\frac{
\mathcal{G}(\mathbf{I})
-
\mu\!\left(\mathcal{G}(\mathbf{I})\right)
}{\tau}
\right),
\label{eq:soft_edge}
\end{equation}
where $\mu(\cdot)$ denotes spatial averaging,
$\sigma(\cdot)$ is the sigmoid function, and $\tau$ is a temperature
parameter.

Based on the soft edge map, we approximate a truncated distance-to-edge
map through iterative min-pooling. It is initialized as
\begin{equation}
\mathcal{D}^{(0)}(\mathbf{I})
=
T\left(1-\mathcal{E}(\mathbf{I})\right),
\label{eq:distance_init}
\end{equation}
and iteratively updated by
\begin{equation}
\mathcal{D}^{(t+1)}(\mathbf{I})
=
\min
\left(
\mathcal{D}^{(t)}(\mathbf{I}),
\operatorname{MinPool}_{3\times3}
\left(
\mathcal{D}^{(t)}(\mathbf{I})
\right)
+1
\right),
\quad
t=0,\ldots,T-1,
\label{eq:distance_update}
\end{equation}
where the minimum is computed element-wise and
$\operatorname{MinPool}_{3\times3}(\cdot)$ computes the local minimum
within a $3\times3$ neighborhood. The final distance-to-edge map is
\begin{equation}
\mathcal{D}(\mathbf{I})
=
\mathcal{D}^{(T)}(\mathbf{I}).
\label{eq:distance_final}
\end{equation}

The structure-matching loss is defined as

\begin{equation}
\mathcal{L}_{\mathrm{struct}}
=\frac{1}{HW}
\left\|
\mathcal{D}(\hat{\mathbf{I}})
-
\mathcal{D}(\mathbf{I})
\right\|_1 + \beta
\frac{1}{HW}
\left\|
\mathcal{E}(\hat{\mathbf{I}})
-
\mathcal{E}(\mathbf{I})
\right\|_1.
\label{eq:structure_loss}
\end{equation}

For the structure-aware variant, the reconstruction objective becomes
\begin{equation}
\mathcal{L}_{\mathrm{rec}}
=
\mathcal{L}_{\mathrm{pix}}
+
\alpha\mathcal{L}_{\mathrm{struct}},
\label{eq:structure_aware_reconstruction}
\end{equation}
where $\alpha$ controls the contribution of structural supervision.
Unless otherwise specified, all reported results use the MSE-only
objective, i.e.,
$\mathcal{L}_{\mathrm{rec}}=\mathcal{L}_{\mathrm{pix}}$.
The structure-aware objective in Eq.~\eqref{eq:structure_aware_reconstruction}
is evaluated only in the document understanding experiments.

\paragraph{Text Generation Objective.}
Meanwhile, the same visual tokens are provided to the text decoder $D_t$ together with task-specific prompts. The decoder autoregressively predicts the textual content depicted in the input image, thereby aligning the visual representations with their corresponding text. We optimize
text generation using the standard autoregressive cross-entropy loss, denoted by $\mathcal{L}_{\mathrm{text}}$.

The overall pre-training objective combines text generation and image reconstruction:
\begin{equation}
\mathcal{L}_{\mathrm{pretrain}}
=
\mathcal{L}_{\mathrm{text}}
+
\lambda \mathcal{L}_{\mathrm{rec}},
\label{eq:pretraining_objective}
\end{equation}
where $\lambda$ balances the two objectives.

Joint optimization provides complementary supervision. Text generation aligns the visual representation with textual content, while image reconstruction encourages the encoder to preserve character strokes, glyph structures, layout information, and other fine-grained visual
evidence that may be discarded under textual supervision alone. Together, these objectives promote more visually grounded representations and improve robustness when linguistic context is weak or unavailable, as further examined through the controlled scrambled-text study in Sec.~\ref{subsec:alpd}.

\textbf{Architecture.} 
We train three vision encoder variants under an identical pre-training objective and on the same MonkeyDoc v2 corpus, differing only in backbone instantiation: ViT-Small, ViT-Base, and ViTAEv2-Small~\cite{vitaev2}, referred to as MonkeyOCRv2-S (28M parameters), MonkeyOCRv2-B (113M parameters), and MonkeyOCRv2-AS (21M parameters), respectively. 
The ViTAEv2-Small variant is adopted for tasks that benefit from its multi-scale inductive bias across resolutions, such as detection, segmentation, and tampering localization. All three encoders are trained from scratch and share the same dual-objective recipe; they constitute a family rather than a single set of weights, and downstream systems load the variant matching their resolution and granularity needs.

\textbf{Training Details.}
We train our models on 64 NVIDIA A800 GPUs. During training, random data augmentation is applied to the input images. The peak learning rate is set to $1\times10^{-3}$ with a global batch size of 256. The reconstruction loss weights $\alpha$, $\beta$, and $\lambda$ are fixed at 0.5, 0.25, and 1.0, respectively, while $T$ and $\tau$ are fixed at 16 and 0.08, without additional hyperparameter tuning. For MonkeyOCRv2-S and MonkeyOCRv2-B, we set the maximum number of input pixels to 1003520 and use a patch size of 14. For MonkeyOCRv2-AS, we use a larger maximum pixel budget of 1802240 with a patch size of 16. We adopt a dynamic resolution training strategy, where the number of visual tokens is adaptively determined according to the input image resolution, following the variable-length tokenization paradigm used in recent VLMs~\cite{qwen25}.

\begin{table*}[t]
\centering
\setlength{\tabcolsep}{4pt}

\caption{For text recognition, integrating MonkeyOCRv2 into the representative CRNN and the leading PARSeq models consistently improves performance across English text (Union14M-Benchmark), Chinese text (Chinese Benchmark), and occluded scene text benchmarks. PARSeq with MonkeyOCRv2 achieves state-of-the-art performance on all three benchmarks. Overall denotes the average performance across the three benchmarks. All results are cited from SVTRv2 except MonkeyOCRv2.
}
\label{tab:comprehensive_results}
\resizebox{\textwidth}{!}{
\begin{tabular}{l cccccccc c ccccc c}
\toprule
\multirow{3}{*}{Model}
&
\multirow{3}{*}{Overall}
&
\multicolumn{8}{c}{Union14M-Benchmark~\cite{union14m}}
&
\multicolumn{5}{c}{Chinese Benchmark~\cite{chinesetext}}
&
\multirow{3}{*}{\makecell[c]{Occlusion Scene\\ Text~\cite{ost}}}
\\
\cmidrule(lr){3-10}
\cmidrule(lr){11-15}
&
&
\textbf{Avg}
&
Artistic
&
\makecell[c]{Contextless}
&
Curve
&
General
&
\makecell[c]{Multi Oriented}
&
\makecell[c]{Multi Words}
&
Saliency
&
\textbf{Avg}
&
Scene
&
Web
&
\makecell[c]{Document}
&
\makecell[c]{Handwriting}
&
\\
\midrule

ABINet~\cite{abinet}
& 73.7
& 75.7 & 71.7 & 74.7 & 80.4 & 79.8 & 69.0 & 76.8 & 77.6
& 70.3 & 66.6 & 63.2 & 98.2 & 53.1
& 75.0
\\

MAERec~\cite{union14m}
& 81.6
& 85.2 & 79.0 & 84.2 & 89.1 & 84.6 & 87.1 & 85.9 & 86.3
& 83.1 & 84.4 & 83.0 & \textbf{99.5} & 65.6
& 76.4
\\

CPPD~\cite{cpdd}
& 81.1
& 81.9 & 76.5 & 82.9 & 86.2 & 83.5 & 78.7 & 81.9 & 83.5
& 81.7 & 82.7 & 82.4 & 99.4 & 62.3
& 79.6
\\

IGTR-AR~\cite{igtrar}
& 81.0
& 84.9 & 77.0 & 82.4 & 90.4 & 84.4 & 91.2 & 84.0 & 84.7
& 81.7 & 82.0 & 81.7 & \textbf{99.5} & 63.8
& 76.3
\\

SMTR~\cite{smtr}
& 80.4
& 85.0 & 76.8 & 83.9 & 89.1 & 83.7 & 87.7 & \textbf{89.3} & 84.6
& 82.7 & 83.4 & 83.0 & 99.3 & 65.1
& 73.5
\\

SVTRv2~\cite{svtrv2}
& 83.1
& 86.1 & \textbf{79.3} & 86.1 & 90.6 & 85.1 & 89.0 & 86.7 & 86.2
& 83.3 & 83.5 & \textbf{83.3} & \textbf{99.5} & 67.0
& 80.0
\\

\midrule

CRNN~\cite{crnn} (ResNet~\cite{resnet})
& 58.7
& 49.2 & 51.2 & 62.3 & 48.1 & 68.2 & 13.0 & 60.4 & 41.4
& 68.8 & 63.8 & 68.2 & 97.0 & 46.1
& 58.0
\\

CRNN (MonkeyOCRv2-S)
& 67.3
& 65.2 & 63.7 & 73.0 & 71.1 & 74.5 & 28.6 & 72.1 & 73.4
& 74.2 & 73.0 & 74.9 & 96.9 & 51.8
& 62.4
\\

\midrule

PARSeq~\cite{parseq} (ViT~\cite{vit})
& 82.2
& 84.3 & 76.5 & 83.4 & 87.6 & 84.9 & 88.8 & 84.3 & 84.4
& 82.4 & 84.2 & 82.8 & \textbf{99.5} & 63.0
& 79.9
\\

PARSeq (MonkeyOCRv2-S)
& \textbf{84.3}
& \textbf{87.6} & 78.6 & \textbf{86.4} & \textbf{92.1} & \textbf{85.4} & \textbf{93.9} & 88.7 & \textbf{87.7}
& \textbf{83.7} & \textbf{84.6} & 83.2 & \textbf{99.5} & \textbf{67.3}
& \textbf{81.5}
\\

\bottomrule
\end{tabular}
}
\end{table*}

\section{Evaluation on Downstream Tasks}

To validate the effectiveness of MonkeyOCRv2, we first evaluate it on five representative document-related tasks: 
text recognition, formula recognition, text detection, document tampering detection, and overlapping text segmentation.
Simply replacing the original visual encoder with MonkeyOCRv2 consistently improves performance across all tasks. We further demonstrate its effectiveness on two more challenging tasks, document parsing and document understanding.
Across all downstream tasks, only the pretrained vision encoder $E_v$ is transferred; the vision decoder and text decoder used during pre-training are discarded. For document parsing and document understanding the encoder is kept frozen, isolating the contribution of the pretrained representation, whereas for the remaining five tasks it is fine-tuned under each system's original protocol.
Throughout this paper, all reported performance differences are absolute
rather than relative and are expressed in percentage points (written with
the \% sign for brevity). When averaging over benchmarks with heterogeneous scales, OCRBench~\cite{ocrbench} is normalized to a 0--100 range by dividing by 10, and thus no single benchmark dominates the mean. Unless otherwise specified, all results except those involving MonkeyOCRv2 are taken directly from the original papers or the official repositories.

\begin{table*}[t]
\centering
\setlength{\tabcolsep}{4pt}
\caption{For formula recognition, integrating MonkeyOCRv2 into UniMERNet-T yields consistent performance gains across three benchmarks: OmniDocBench 1.6 for formulas from diverse document types, MathWriting for irregular handwritten expressions, and UniMER-Test with four subsets covering printed, handwritten and screen-capture scenarios. ExpRate denotes the percentage of expressions exactly matching the ground truth. The results of Pix2tex, Texify and UniMERNet on UniMER-Test are cited from UniMERNet.}
\label{tab:unimer_test_results}
\resizebox{\textwidth}{!}{
\begin{threeparttable}
\begin{tabular}{l c cc cc cc cc cc cc cc}
\toprule
\multirow{2}{*}{Model} &
\multirow{2}{*}{Params} &
\multicolumn{2}{c}{Overall} &
\multicolumn{2}{c}{OmniDocBench 1.6~\cite{omnidocbench}} &
\multicolumn{2}{c}{MathWriting~\cite{mathwriting}} &
\multicolumn{2}{c}{\makecell[c]{Simple Printed\\Expressions~\cite{unimernet}}} &
\multicolumn{2}{c}{\makecell[c]{Complex Printed\\ Expressions~\cite{unimernet}}} &
\multicolumn{2}{c}{\makecell[c]{Handwritten\\ Expressions~\cite{unimernet}}} &
\multicolumn{2}{c}{\makecell[c]{Screen Capture\\ Expressions~\cite{unimernet}}} \\
\cmidrule(lr){3-4} \cmidrule(lr){5-6} \cmidrule(lr){7-8}
\cmidrule(lr){9-10} \cmidrule(lr){11-12} \cmidrule(lr){13-14}
\cmidrule(lr){15-16}
& & 
CDM & ExpRate &
CDM & ExpRate &
CDM & ExpRate &
CDM & ExpRate &
CDM & ExpRate &
CDM & ExpRate &
CDM & ExpRate \\
\midrule
Pix2tex\tnote{1}     & 25.5M  & 53.8 & 23.3 & 69.4 & 27.0 & 0.4 & 0.0 & 96.2 & 72.4 & 64.9 & 7.1 & 24.5 & 0.6 & 67.6 & 32.8 \\
Texify\tnote{2}     & 312M  & 67.3 & 40.4 & 76.5 & 46.4 & 26.6 & 2.0 & 98.5 & 91.0 & 70.4 & 28.2 & 52.7 & 23.6 & 79.3 & 51.3 \\
UniMERNet-B~\cite{unimernet} & 325M & 89.5 & 64.5 & 90.4 & 59.5 & 63.8 & 12.3 & 99.1 & 93.3 & 96.0 & \textbf{80.5} & 94.0 & 64.3 & 93.7 & 77.0 \\
UniMERNet-S~\cite{unimernet} & 202M & 89.8 & 63.9 & 90.1 & 59.1 & 65.9 & 12.7 & 99.1 & 93.4 & 95.9 & 77.7 & 93.7 & 63.9 & \textbf{94.1} & 76.9 \\

\midrule
UniMERNet-T~\cite{unimernet} (Swin~\cite{swin}) & 107M & 89.4 & 61.8 & 89.9 & 57.2 & 65.6 & 12.9 & 99.1 & 92.3 & 94.9 & 69.9 & 93.3 & 61.9 & 93.8 & 76.6 \\
UniMERNet-T (MonkeyOCRv2-S) & 110M & \textbf{90.9} & \textbf{66.4} & \textbf{90.8} & \textbf{61.1} & \textbf{70.8} & \textbf{16.2} & \textbf{99.2} & \textbf{93.8} & \textbf{96.1} & 79.2 & \textbf{94.3} & \textbf{69.5} & 94.0 & \textbf{78.6} \\
\bottomrule
\end{tabular}

\begin{tablenotes}[para,flushleft]
\footnotesize
\setlength{\itemsep}{0pt}
\setlength{\parskip}{0pt}
\raggedright
\item[1] Pix2tex: https://github.com/lukas-blecher/LaTeX-OCR
\item[2] Texify: https://github.com/VikParuchuri/texify
\end{tablenotes}

\end{threeparttable}
}
\end{table*}

\subsection{Text Recognition}
    Text recognition aims to recognize the textual content in images. 
    We adopt the representative CRNN~\cite{crnn} and the leading PARSeq~\cite{parseq}, and replace their original visual encoders with MonkeyOCRv2-S, which is comparable in parameter count to the encoders it replaces, keeping the rest unchanged.
    Following the protocol of SVTRv2~\cite{svtrv2}, we train the models separately on Union14M~\cite{union14m} and a Chinese text recognition dataset~\cite{chinesetext}.

    \textbf{Datasets.} Following SVTRv2, we evaluate our model on three benchmarks: (1) Union14M-Benchmark~\cite{union14m}, comprising seven challenging subsets: Curve, Multi-Oriented, Artistic, Contextless, Salient, Multi-Words, and General; (2) Chinese benchmark~\cite{chinesetext}, consisting of Scene, Web, Document, and Handwriting subsets; and  (3) the occluded scene text benchmark~\cite{ost} (OST), containing both weakly and heavily occluded scene text images.
    We also evaluate our model on the common benchmarks (IC13~\cite{ic13}, SVT~\cite{svt}, IIIT5K~\cite{IIIT5K}, IC15~\cite{ic15}, SVTP~\cite{SVTP}, and CUTE80~\cite{cute80}). PARSeq with MonkeyOCRv2 achieves an average accuracy of 96.8\% across the six datasets, surpassing the previous state-of-the-art SVTRv2 (96.6\%), as shown in App.~\ref{sec:appendix_common}. However, since performance on these benchmarks is largely saturated, we focus our analysis on the three more challenging benchmarks described above.

    \textbf{Results.} As shown in Tab.~\ref{tab:comprehensive_results}, MonkeyOCRv2 consistently improves both CRNN and PARSeq across all benchmarks. On the challenging English Union14M benchmark, replacing the visual encoder with MonkeyOCRv2 brings a 16.0\% absolute average accuracy gain for CRNN and a 3.3\% gain for PARSeq, which reaches 87.6\% accuracy, 1.5\% above the previous best SVTRv2 (86.1\%). On the multi-scene Chinese benchmark, MonkeyOCRv2 demonstrates strong cross-domain generalization, improving CRNN and PARSeq by 5.4\% and 1.3\%  on average, respectively. It also improves performance on the occluded scene text benchmark, yielding 4.4\% and 1.6\% gains for CRNN and PARSeq. Overall, MonkeyOCRv2 consistently improves recognition across diverse text scenarios, and PARSeq equipped with MonkeyOCRv2 achieves the best overall performance, outperforming previous state-of-the-art methods.

\subsection{Formula Recognition}
    Formula recognition aims to convert formula images into structured LaTeX sequences. Unlike text recognition, it is more challenging as it requires capturing complex spatial relationships, such as the vertical arrangement between formula symbols. We build upon the widely used UniMERNet-T~\cite{unimernet}, replacing its original vision encoder pretrained on 16M in-house data with MonkeyOCRv2-S, while keeping the rest of the model architecture and training configuration unchanged.

    \textbf{Datasets.} We follow the training pipeline and CDM evaluation metric~\cite{cdm} of UniMERNet~\cite{unimernet}. Since UniMERNet leverages 16M unreleased in-house samples for additional pre-training, we exclude this closed pre-training stage and only adopt its public fine-tuning configuration on the UniMER-1M dataset.  We conduct evaluations on three widely used mathematical expression recognition benchmarks: (1) OmniDocBench 1.6~\cite{omnidocbench}: A multi-scene document benchmark collected from various PDF files, annotated with LaTeX labels for embedded formula regions. We crop formula regions via the provided bounding boxes for evaluation. (2) MathWriting~\cite{mathwriting}: A handwritten mathematical expression benchmark consisting of real handwritten and synthetic formula samples, targeting irregular handwritten formula recognition. (3) UniMER-Test~\cite{unimernet}: The official test set of the UniMER suite for real-world mathematical expression recognition. It contains four fine-grained subsets 
    with differing difficulty levels and scenarios: Simple Printed Expressions, Complex Printed Expressions, Handwritten Expressions, and Screen Capture Expressions. 
    
    \textbf{Results.} As shown in Tab.~\ref{tab:unimer_test_results}, replacing UniMERNet-T's original Swin Transformer encoder (pretrained on their 16M in-house data) with MonkeyOCRv2 yields consistent gains across benchmarks. On OmniDocBench 1.6, which evaluates formulas from diverse document types, MonkeyOCRv2 improves CDM by 0.9\% and ExpRate by 3.9\% absolute, demonstrating strong generalization on real-world document formula scenarios. On MathWriting, the benchmark for irregular handwritten expressions, MonkeyOCRv2 brings more notable improvements: CDM increases by 5.2\% and ExpRate by 3.3\%, reflecting its enhanced ability to capture fine-grained text features. On 
    UniMER-Test, MonkeyOCRv2 achieves consistent improvements across four subsets, with particularly large gains on challenging scenarios such as Complex Printed Expressions and Handwritten Expressions, where ExpRate improves by 9.3\% and 7.6\%, respectively. Overall, with only 110M parameters, our model surpasses the larger 325M UniMERNet-B, 
    demonstrating that a stronger document-oriented visual encoder can compensate for reduced model capacity.

\begin{figure*}[t]
    \centering
    \includegraphics[width=\textwidth]{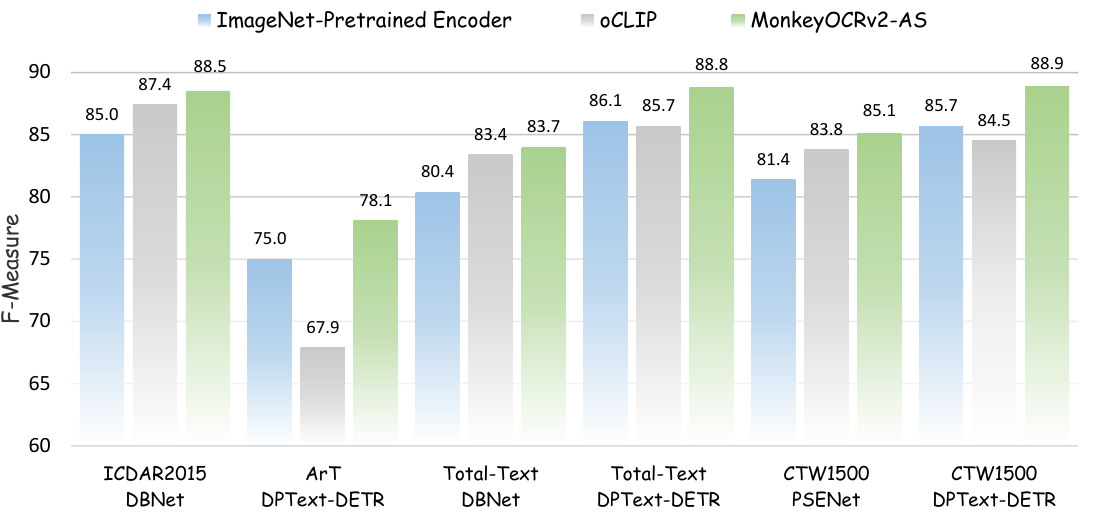}
    \caption{For text detection, MonkeyOCRv2 consistently delivers robust performance gains on ICDAR2015~\cite{ic15}, ArT~\cite{chng2019icdar2019}, Total-Text~\cite{ch2020total}, and CTW1500~\cite{liu2019curved} with DBNet~\cite{dbnet}, PSENet~\cite{wang2019shape}, and DPText-DETR~\cite{dptextdetr}. It outperforms both the original ImageNet-pretrained encoder and oCLIP~\cite{oclip}, a visual encoder pretrained for text detection. All results are reproduced by us.}
\label{fig:text_detection}
\end{figure*}

\subsection{Text Detection}

    Text detection aims to localize textual regions in unconstrained natural scene images. Since the task requires multi-scale feature representation to handle texts with varying sizes, we replace the original visual encoders of baseline methods with MonkeyOCRv2-AS.
    
    \textbf{Datasets.} We evaluate on four widely used scene text detection benchmarks: (1) ICDAR 2015~\cite{ic15}: A scene text benchmark featuring multi-oriented and distorted text captured under unconstrained real-world conditions; (2) ArT~\cite{chng2019icdar2019}: An arbitrary-shaped scene text benchmark featuring diverse text shapes, including horizontal, multi-oriented, and curved text. (3) Total-Text~\cite{ch2020total}: A curved scene text benchmark containing horizontal, multi-oriented, and curved text instances, designed to evaluate models on diverse text orientations and irregular text shapes; (4) CTW1500~\cite{liu2019curved}: A curved scene text benchmark covering diverse scenarios, including indoor/outdoor scenes, blurred and perspective-distorted text, and multilingual text.
    
    \textbf{Training Settings.}
    We verify the effectiveness of our encoder on two representative detection models, DBNet~\cite{dbnet} and PSENet~\cite{wang2019shape}, as well as the leading DPText-DETR~\cite{dptextdetr} model. For each detector, we compare three visual backbones under identical settings: the original ImageNet-pretrained encoder;    oCLIP~\cite{oclip}, a visual encoder pretrained for text detection; and MonkeyOCRv2. To ensure fairness, all models are fine-tuned directly on the target detection datasets without any additional text detection pre-training.

\begin{table*}[t]
\centering
\caption{Effectiveness of MonkeyOCRv2 on document tampering detection. * denotes models trained by us using the ViTAEv2~\cite{vitaev2} visual encoder pretrained with DeepSolo~\cite{deepsolo}, configured to have a comparable number of parameters to MonkeyOCRv2-AS. All other results are cited from FFDN.}
\resizebox{\textwidth}{!}{
\begin{tabular}{ll|cc|cccc|cccc|cccc}
\toprule
\multirow{2}{*}{Method} & \multirow{2}{*}{Param.}
& \multicolumn{2}{c}{Overall}
& \multicolumn{4}{c}{DocTamper-Test} 
& \multicolumn{4}{c}{DocTamper-FCD} 
& \multicolumn{4}{c}{DocTamper-SCD} \\
\cmidrule(lr){3-4}\cmidrule(lr){5-8}\cmidrule(lr){9-12}\cmidrule(lr){13-16}
& & IoU & F & IoU & P & R & F 
& IoU & P & R & F
& IoU & P & R & F \\
\midrule
PSCC-Net~\cite{pscc-net} & 5M
& 13.7 & 31.3
& 17.0 & 25.0 & 83.0 & 39.0
& 13.0 & 19.0 & 82.0 & 30.0
& 11.0 & 15.0 & \textbf{83.0} & 25.0 \\
UperNet~\cite{upernet} & 67M
& 49.3 & 54.0
& 70.0 & 66.0 & 60.0 & 62.0
& 30.0 & 57.0 & 35.0 & 43.0
& 48.0 & 57.0 & 58.0 & 57.0 \\
CAT-Net~\cite{catnet} & 114M
& 67.3 & 71.0
& 78.0 & 75.0 & 69.0 & 72.0
& 66.0 & 85.0 & 70.0 & 76.0
& 58.0 & 65.0 & 65.0 & 65.0 \\
Swin-UPer~\cite{swin} & 81M
& 66.7 & 71.7
& 79.0 & 75.0 & 72.0 & 73.0
& 64.0 & 80.0 & 70.0 & 75.0
& 57.0 & 66.0 & 68.0 & 67.0 \\
SegFormer~\cite{segformer} & 85M
& 70.3 & 74.0
& 81.0 & 77.0 & 74.0 & 75.0
& 69.0 & 82.0 & 74.0 & 78.0
& 61.0 & 68.0 & 70.0 & 69.0 \\
Mask2Former~\cite{mask2former} & 69M
& 69.7 & 78.0
& 84.0 & 82.0 & 83.0 & 82.0
& 66.0 & 81.0 & 75.0 & 78.0
& 59.0 & 70.0 & 79.0 & 74.0 \\
ConvNeXt~\cite{convnext} & 122M
& 69.7 & 75.3
& 84.0 & 81.0 & 78.0 & 79.0
& 62.0 & 76.0 & 71.0 & 74.0
& 63.0 & 71.0 & 74.0 & 73.0 \\
ConvNeXtV2~\cite{convnextv2} & 121M
& 72.7 & 77.7
& 86.0 & 82.0 & 79.0 & 81.0
& 65.0 & 79.0 & 75.0 & 77.0
& 67.0 & 74.0 & 76.0 & 75.0 \\
InternImage~\cite{internimage} & 128M
& 73.3 & 77.7
& 84.0 & 81.0 & 77.0 & 79.0
& 72.0 & 83.0 & 79.0 & 81.0
& 64.0 & 73.0 & 74.0 & 73.0 \\
ASC-Former~\cite{luo25rtm} & 80M
& 68.2 & 80.8
& 81.5 & 91.8 & 87.8 & 89.8
& 61.3 & 74.9 & 77.1 & 76.0
& 61.9 & 78.0 & 75.0 & 76.5 \\
DTD~\cite{qu2023towards} & 66M
& \underline{77.0} & 79.7
& 84.0 & 81.0 & 77.0 & 79.0
& 79.0 & 88.0 & 82.0 & 85.0
& \textbf{68.0} & 75.0 & 76.0 & 75.0 \\

\midrule

FFDN*~\cite{chen2024enhancing} (ViTAEv2~\cite{vitaev2}) & 69M
& 70.7 & \underline{82.7}
& 69.4 & 76.2 &  88.7 & 82.0
& 79.0 & \textbf{92.5} & 84.4 & 88.3
& 63.6 & 79.1 & 76.5 & 77.8 \\
 
FFDN (MonkeyOCRv2-AS) & 71M
& \textbf{78.2} & \textbf{87.5}
& \textbf{87.4} & \textbf{94.8} & \textbf{91.8} & \textbf{93.3}
& \textbf{79.9} & 90.4 & \textbf{87.4} & \textbf{88.9}
& 67.2 & \textbf{81.0} & 79.8 & \textbf{80.4} \\

\bottomrule
\end{tabular}
}
\label{tab:doctamper_results}
\end{table*}

    \textbf{Results.} Fig.~\ref{fig:text_detection} summarizes the detection results on four benchmarks. MonkeyOCRv2 consistently improves F-measure across all datasets and detector architectures, demonstrating stronger compatibility and robustness than both the ImageNet-pretrained encoder and the text-specific oCLIP encoder. On the ICDAR 2015 benchmark, replacing the ImageNet-pretrained encoder with MonkeyOCRv2 in DBNet improves F-measure from 85.0 to 88.5, and further surpasses oCLIP by 1.1\%, highlighting the superior feature representation of MonkeyOCRv2. On the challenging ArT arbitrary-shaped text benchmark, MonkeyOCRv2 improves DPText-DETR by 3.1\% in F-measure, while oCLIP shows limited compatibility with this modern DETR-based detector, further demonstrating the robustness of MonkeyOCRv2. On the Total-Text curved text benchmark, MonkeyOCRv2 improves DBNet by 3.3\% over the ImageNet-pretrained encoder. For the stronger DPText-DETR, oCLIP fails to provide improvements, whereas MonkeyOCRv2 achieves a 2.7\% F-measure gain. On CTW1500, MonkeyOCRv2 boosts F-measure by 3.7\% and 3.2\% for PSENet and DPText-DETR, respectively. 
    Overall, MonkeyOCRv2 consistently delivers robust improvements across diverse datasets and architectures, whereas the gains of oCLIP fail to transfer to the modern DETR-based detector.

\subsection{Overlapping Text Segmentation}
    Overlapping text segmentation aims to predict pixel masks for the occluding text, the occluded text, and their overlap region. Since the task requires fine-grained multi-scale feature representation to distinguish text boundaries and handle occlusions of varying scales, we replace the original visual backbones of baseline models with MonkeyOCRv2-AS.
    
    \textbf{Datasets.} We evaluate on the MOT dataset~\cite{liu2025multi}, which contains Chinese and English overlapping-text samples from diverse real-world scenarios, including printed documents, receipts, artistic text, and doorplates.

\begin{wraptable}[16]{r}{0.55\linewidth}
\vspace{-11pt}
    \centering
    \setlength{\tabcolsep}{2.5pt}
    \caption{
    Effectiveness of MonkeyOCRv2 on overlapping text segmentation. mIoU\textsubscript{Text} denotes 
    the mean IoU over the occlusion, occluded, and overlap regions. All results are cited from MOTS except MonkeyOCRv2. 
    }
    \label{tab:overlap_results}
    \resizebox{\linewidth}{!}{
    \begin{tabular}{l c c c c}
        \toprule
        Model & \textbf{mIoU\textsubscript{Text}} & IoU\textsubscript{Occlusion} & IoU\textsubscript{Occluded} & IoU\textsubscript{Overlap} \\
        \midrule
        Unet~\cite{unet}                  & 62.2 & 80.2 & 65.7 & 40.7 \\
        Deeplab v3~\cite{deeplabv3+}      & 67.9 & 83.2 & 71.2 & 49.3 \\
        OCRNet~\cite{yuan2020object}      & 65.8 & 81.0 & 68.5 & 47.8 \\
        Segformer~\cite{segformer}        & 69.0 & 83.6 & 74.1 & 49.3 \\
        MaskFormer~\cite{cheng2021per}    & 68.4 & 83.5 & 70.3 & 51.4 \\
        TexRNet~\cite{xu2021rethinking}   & 68.9 & 84.2 & 73.2 & 49.3 \\
        EAFormer~\cite{yu2024eaformer}    & 69.1 & 83.8 & 74.2 & 50.5 \\
        WASNet~\cite{xie2024dataset}      & 70.8 & 84.8 & 74.4 & 53.1 \\
        \midrule
        Mask2Former~\cite{mask2former} (ResNet)    & 70.3 & 84.7 & 73.3 & 52.8 \\
        Mask2Former (MonkeyOCRv2-AS)                     & \underline{76.6} & \textbf{88.6} & \textbf{83.4} & \underline{57.7} \\
        \midrule
        MOTS~\cite{liu2025multi} (ResNet)         & 72.6 & 85.2 & 77.5 & 54.9 \\
        MOTS (MonkeyOCRv2-AS)                 & \textbf{76.9} & \textbf{88.6} & \underline{82.6} & \textbf{59.4} \\
        \bottomrule
    \end{tabular}}
\end{wraptable}
    
    \textbf{Training Details.} 
    We integrate MonkeyOCRv2 into the widely used Mask2Former~\cite{mask2former} and MOTS~\cite{liu2025multi}.
    For each model, we replace the original backbone with MonkeyOCRv2 of comparable parameter scale, while keeping all other configurations and hyperparameters identical to the original papers for fair comparison.

    \textbf{Results.} As shown in Tab.~\ref{tab:overlap_results}, MonkeyOCRv2 consistently improves performance across all metrics for both baseline models. For 
    Mask2Former, the mIoU\textsubscript{Text} rises from 70.3\% to 76.6\%, a 6.3\% gain. 
    For MOTS,
    the mIoU\textsubscript{Text} increases from 72.6\% to 76.9\%, with a 4.3\% improvement. Our model achieves the best performance among all compared methods, demonstrating the effectiveness of MonkeyOCRv2 as a visual backbone for overlapping text segmentation.

\subsection{Document Tampering Detection}
    Document tampering detection aims to localize manipulated regions in document images. Since the task requires fine-grained artifact perception to identify subtle tampering traces, we replace the original visual encoder of the baseline model with MonkeyOCRv2-AS.

    \textbf{Datasets.} We conduct experiments on the DocTamper benchmark~\cite{qu2023towards}, which consists of one standard test set and two cross-domain test sets: (1) DocTamper-Test: 
    The standard test set, whose data distribution and document styles match the training set, for in-domain evaluation;
    (2) DocTamper-FCD: First cross-domain test subset with document textures and layouts 
    entirely different from the training data
    to verify generalization; (3) DocTamper-SCD: Second cross-domain test subset with document scenes and styles greatly divergent from training samples for strict generalization testing.
    
    \textbf{Training Settings.} We adopt the state-of-the-art FFDN~\cite{chen2024enhancing} as the baseline framework. To fairly compare the representation capacity of different backbones, we follow the original FFDN training protocol and replace its visual encoder with two text-centric encoders of comparable parameters: the DeepSolo-pretrained ViTAEv2~\cite{deepsolo} and our MonkeyOCRv2. All experiments follow the official DocTamper evaluation protocol, and we report IoU, Precision, Recall, and F1-score as evaluation metrics.
    
    \textbf{Results.} As shown in Tab.~\ref{tab:doctamper_results}, replacing the visual encoder of FFDN with MonkeyOCRv2 consistently improves performance across all three test sets, outperforming the DeepSolo-pretrained ViTAEv2 backbone and achieving state-of-the-art overall results. On the standard DocTamper-Test set, our method achieves 87.4\% IoU and 93.3\% F1-score, with an 11.3\% absolute F1 gain over the baseline, demonstrating strong fine-grained tampering localization ability. On the DocTamper-FCD cross-domain set, MonkeyOCRv2 reaches the best F1-score of 88.9\%, maintaining robust performance under font and format domain shifts. On the DocTamper-SCD cross-domain set, our method also obtains the highest F1-score of 80.4\%, verifying favorable generalization capability across scene-level domain discrepancies. These results confirm that MonkeyOCRv2 provides high-quality document visual representations, and delivers stable improvements for both in-domain and cross-domain document tampering detection.

\begin{table*}[tbp]
\centering

\caption{
Performance comparison on MDPBench~\cite{mdpbench}, a challenging multilingual document parsing benchmark.
MonkeyOCRv2-B-Parsing outperforms all other open-source models, with its vision encoder having only $1/11$ the parameters of the previous state-of-the-art dots.mocr. All results are under the official MDPBench evaluation protocol.
}
\label{tab:evaluation_results_combined_all}

\resizebox{0.9\linewidth}{!}{
\begin{threeparttable}
\begin{tabular}{l c c c c c c c c}
\toprule
\textbf{Model} & \textbf{\makecell{Total\\Params}} & \textbf{\makecell{ViT\\Params}} & \textbf{\makecell{LLM\\Params}} & \textbf{All} & \textbf{Digit.} & \textbf{Photo.} & \textbf{Latin} & \textbf{Non-Latin} \\
\midrule
\multicolumn{9}{l}{\textbf{\textit{Closed-source VLMs}}} \\
\midrule
GPT-5.2\tnote{1} & -- & -- & -- & 68.6 & 85.6 & 63.0 & 75.2 & 61.1 \\
Claude-Sonnet-4.6\tnote{2} & -- & -- & -- & 73.1 & 85.0 & 69.3 & 79.2 & 66.2 \\
Doubao-2.0-pro\tnote{3} & -- & -- & -- & 74.2 & 78.9 & 72.8 & 75.7 & 72.5 \\
Gemini-3-pro\tnote{4} & -- & -- & -- & \textbf{86.4} & \textbf{90.4} & \textbf{85.1} & \textbf{88.4} & \textbf{84.1} \\
\midrule
\multicolumn{9}{l}{\textbf{\textit{Open-source VLMs}}} \\
\midrule
InternVL-3.5-8B~\cite{internvl3.5} & 8.3B & 0.3B & 8B & 42.7 & 59.7 & 37.0 & 53.4 & 30.6 \\
MinerU-2.5~\cite{mineru25} & 1.2B & 0.7B & 0.5B & 46.3 & 61.9 & 40.8 & 63.0 & 27.4 \\
DeepSeek-OCR~\cite{deepseekocr} & 3.4B & 0.4B & 3B & 51.8 & 80.7 & 42.2 & 54.5 & 48.9 \\
MonkeyOCR-pro-3B~\cite{monkeyocr} & 3.7B & 0.7B & 3B & 52.2 & 68.0 & 47.0 & 65.1 & 37.6 \\
Nanonets-OCR-s\tnote{5} & 3.7B & 0.7B & 3B & 63.7 & 78.8 & 58.7 & 71.3 & 55.0 \\
Nanonets-OCR2-3B\tnote{6} & 3.7B & 0.7B & 3B & 64.2 & 79.2 & 59.3 & 71.4 & 56.2 \\
Qwen3.5-Instruct-9B\tnote{7} & 9.7B & 0.7B & 9B & 65.7 & 74.8 & 62.7 & 72.5 & 58.2 \\
GLM-OCR~\cite{glm-ocr} & 0.9B & 0.4B & 0.5B & 67.3 & 77.9 & 63.7 & 78.7 & 54.3 \\
Qwen3-VL-Instruct-8B~\cite{qwen3vl} & 8.3B & 0.3B & 8B & 68.3 & 78.4 & 65.0 & 73.6 & 62.5 \\
HunyuanOCR~\cite{hunyuanocr} & 1.0B & 0.4B & 0.6B & 68.3 & 80.2 & 64.3 & 72.4 & 63.7 \\
PaddleOCR-VL~\cite{ppocrvl} & 0.9B & 0.6B & 0.3B & 69.6 & 87.6 & 63.6 & 72.1 & 66.7 \\
olmOCR2~\cite{olmocr2} & 7.7B & 0.7B & 7B & 70.4 & 79.9 & 67.2 & 76.7 & 63.3 \\
MinerU-2.5-pro~\cite{mineru2.5pro} & 1.2B & 0.7B & 0.5B & 71.0 & 86.2 & 66.1 & 74.6 & 67.0 \\
PaddleOCR-VL-1.6~\cite{ppocrvl16} & 0.9B & 0.6B & 0.3B & 75.0 & 82.8 & 72.6 & 78.0 & 71.6 \\
HunyuanOCR-1.5~\cite{hunyuanocr15} & 1.0B & 0.4B & 0.6B & 76.8 & 86.2 & 73.6 & 79.7 & 73.5 \\
Kimi-K2.5~\cite{kimi-k2.5} & 1.0T & 0.4B & 1T & 77.5 & 85.0 & 75.0 & 81.6 & 72.9 \\
PaddleOCR-VL-1.5~\cite{ppocrvl15} & 0.9B & 0.6B & 0.3B & 78.3 & 87.4 & 75.2 & 81.2 & 74.9 \\
chandra-ocr-2\tnote{8} & 5.3B & 0.5B & 4.8B & 79.7 & 87.8 & 77.1 & 82.7 & 76.4 \\
dots.mocr~\cite{dotsmocr} & 3.0B & 1.2B & 1.8B & 80.5 & \textbf{90.5} & 77.2 & 81.7 & 79.2 \\
\textbf{MonkeyOCRv2-S-Parsing} & \textbf{0.6B} & \textbf{0.03B} & 0.6B & \underline{82.5} & 87.9 & \underline{80.7} & \underline{83.2} & \underline{81.7} \\
\textbf{MonkeyOCRv2-B-Parsing} & \underline{0.7B} & \underline{0.1B} & 0.6B & \textbf{83.3} & \underline{88.1} & \textbf{81.7} & \textbf{84.2} & \textbf{82.1} \\
\bottomrule
\end{tabular}
\begin{tablenotes}[para,flushleft]
\tiny
\setlength{\itemsep}{0pt}
\setlength{\parskip}{0pt}
\raggedright
\item[1] GPT-5.2: https://chat.openai.com
\item[2] Claude-Sonnet-4.6: https://www.anthropic.com/claude
\item[3] Doubao-2.0-pro: https://research.doubao.com
\item[4] Gemini-3-pro: https://blog.google/innovation-and-ai/technology/developers-tools/gemini-3-pro-vision
\item[5] Nanonets-OCR-s: https://nanonets.com/research/nanonets-ocr-s
\item[6] Nanonets-OCR2-3B: https://nanonets.com/research/nanonets-ocr-2
\item[7] Qwen3.5-Instruct-9B: https://qwen.ai/blog?id=qwen3.5
\item[8] chandra-ocr-2: https://www.datalab.to/blog/chandra-2
\end{tablenotes}
\end{threeparttable}
}
\end{table*}

\begin{table*}[t]
\centering

\caption{Comprehensive evaluation on OmniDocBench 1.6~\cite{omnidocbench}, which contains only Chinese and English documents. MonkeyOCRv2-Parsing achieves competitive performance even with a frozen vision encoder and without any additional task-specific post-training. All results are cited from OmniDocBench 1.6 except MonkeyOCRv2}
\label{tab:omni16_full_performance}

\resizebox{1\textwidth}{!}{
\begin{threeparttable}
\renewcommand{\arraystretch}{1.2}
\begin{tabular}{lcc|c|c c c c c}
\toprule
\multirow{2}{*}{\textbf{Methods}} & \multirow{2}{*}{\textbf{\makecell[c]{Unfreeze\\ViT}}} & \multirow{2}{*}{\textbf{\makecell[c]{Post-\\Training}}} & \multirow{2}{*}{\textbf{Overall$\uparrow$}} & \multirow{2}{*}{\textbf{Text\textsuperscript{Edit}$\downarrow$}} & \multirow{2}{*}{\textbf{Formula\textsuperscript{CDM}$\uparrow$}} & \multirow{2}{*}{\textbf{Table\textsuperscript{TEDS}$\uparrow$}} & \multirow{2}{*}{\textbf{Table\textsuperscript{TEDS-S}$\uparrow$}} & \multirow{2}{*}{\textbf{Reading Order\textsuperscript{Edit}$\downarrow$}} \\
&&&&&&&&\\
\midrule
Nanonets-OCR-s\tnote{1} & \checkmark & \XSolidBrush& 83.61 & 0.108 & 81.46 & 80.18 & 84.51 & 0.213 \\
InternVL3.5-241B~\cite{internvl3.5} & \checkmark & \XSolidBrush & 83.76 & 0.130 & 89.95 & 74.35 & 79.78 & 0.215 \\
Kimi K2.5~\cite{kimi-k2.5} & \checkmark & \XSolidBrush & 84.53 & 0.107 & 83.50 & 80.76 & 84.00 & 0.211 \\
olmOCR~\cite{olmocr} & \checkmark & \XSolidBrush  & 85.74 & 0.139 & 88.10 & 83.00 & 87.17 & 0.216 \\
GPT-5.2\tnote{2} & - & - &  86.59 & 0.114 & 88.21 & 82.95 & 87.93 & 0.193 \\
MonkeyOCR-pro-3B~\cite{monkeyocr} & \XSolidBrush & \XSolidBrush & 88.57 & 0.074 & 88.74 & 84.35 & 88.62 & 0.189 \\
Qwen3-VL-235B~\cite{qwen3vl} & \checkmark & \XSolidBrush & 89.78 & 0.063 & 92.55 & 83.07 & 86.75 & 0.166 \\
HunyuanOCR~\cite{hunyuanocr} & \checkmark & \checkmark & 89.95 & 0.088 & 87.68 & 91.01 & 93.23 & 0.171 \\
DeepSeek-OCR 2~\cite{deepseekocr2} & \checkmark & \XSolidBrush & 90.25 & 0.050 & 91.84 & 83.89 & 87.75 & 0.144 \\
dots.ocr~\cite{dotsocr} & \checkmark & \XSolidBrush & 90.77 & 0.048 & 89.95 & 87.18 & 90.58 & 0.138 \\
Gemini-3-pro\tnote{3} & - & - & 92.91 & 0.064 & 95.99 & 89.15 & 92.96 & 0.165 \\

HunyuanOCR-1.5~\cite{hunyuanocr15} & \checkmark & \checkmark & 94.74 & 0.039  & 94.50 & 93.67 & 94.71 & 0.129 \\

GLM-OCR~\cite{glm-ocr} & \checkmark & \checkmark & 95.22 & 0.044 & 97.18 & 92.83 & 95.39 & 0.133 \\
MinerU2.5-pro~\cite{mineru2.5pro} & \checkmark & \checkmark & \underline{95.75} & \underline{0.036} & \underline{97.45} & \underline{93.42} & \underline{95.92} & \textbf{0.120} \\
PaddleOCR-VL-1.6~\cite{ppocrvl16} & \checkmark & \checkmark & \textbf{96.33} & \textbf{0.033} & \textbf{97.49} & \textbf{94.76} & \textbf{97.11} & \underline{0.127} \\
\midrule
MonkeyOCRv2-S-Parsing & \XSolidBrush & \XSolidBrush & {90.90} & {0.055} & {90.57} & {87.59} & {90.64} & 0.134 \\
MonkeyOCRv2-B-Parsing & \XSolidBrush  & \XSolidBrush & {91.57} & {0.053} & {91.83} & {88.24} & {91.38} & 0.131 \\
\bottomrule
\end{tabular}
\begin{tablenotes}[para,flushleft]
\scriptsize
\setlength{\itemsep}{0pt}
\setlength{\parskip}{0pt}
\raggedright
\item[1] Nanonets-OCR-s: https://nanonets.com/research/nanonets-ocr-s
\item[2] GPT-5.2: https://chat.openai.com
\item[3] Gemini-3-pro: https://blog.google/innovation-and-ai/technology/developers-tools/gemini-3-pro-vision
\end{tablenotes}
\end{threeparttable}
}
\end{table*}
    
\subsection{Document Parsing}
Document parsing aims to systematically convert the complex multimodal content of document images into structured information. Following the prevailing paradigm, we combine our frozen visual encoder with a large language model to build MonkeyOCRv2-Parsing, a 0.7B document parsing model.

\textbf{Architecture.} The architecture comprises three components: a vision encoder, an MLP projector, and a large language model (LLM). 
We instantiate the vision encoder with either MonkeyOCRv2-S or MonkeyOCRv2-B, yielding MonkeyOCRv2-S-Parsing and MonkeyOCRv2-B-Parsing, and adopt Qwen3-0.6B~\cite{qwen3} as the LLM.
Given a document image, MonkeyOCRv2-Parsing first predicts the coordinates and categories of document elements in natural reading order, providing an explicit layout structure for subsequent extraction. Each detected element is then cropped by its predicted bounding box and fed back to MonkeyOCRv2-Parsing  with a task-specific prompt for parallel content recognition. Finally, the recognized elements are assembled according to the predicted reading order into the structured document output.

\textbf{Training Details.} The parsing model is trained in two stages. In the first stage, we train only the MLP projector for vision-language alignment with a learning rate of $2 \times 10^{-4}$. In the second stage, we train the MLP and the LLM jointly with a learning rate of $2 \times 10^{-5}$. The vision encoder remains frozen throughout. We cap the input at $1280$ visual tokens, each corresponding to a $28 \times 28$ pixel patch, set the maximum sequence length to $16{,}384$, and use a global batch size of $256$. The model is trained for one epoch in approximately $6$--$7$ days on $64$ NVIDIA A800 GPUs.

\textbf{Evaluation.} We evaluate primarily on MDPBench~\cite{mdpbench}, which spans both digital-born and photographed documents across 17 languages. We choose it because its joint coverage of photographed inputs and non-Latin scripts represents precisely the regime where fine-grained character-level perception determines accuracy, making it a discriminative testbed for the capability targeted in this paper.
While MDPBench originates from the same research line as our prior benchmarks, the encoder, the LLM, and the data pipeline evaluated here are independent of its construction, and we additionally report on OmniDocBench 1.6 below for cross-benchmark calibration.
As shown in Tab.~\ref{tab:evaluation_results_combined_all}, MonkeyOCRv2-Parsing attains the best result among all open-source models, raising the MDPBench overall score from $80.5$ (the 3B dots.mocr) to $83.3$ and surpassing the 0.9B PaddleOCR-VL-1.6 ($75.0$), while its $0.1$B vision encoder is roughly $5$ and $11$ times smaller than those of PaddleOCR-VL-1.6 and dots.mocr, respectively.
We further report results on OmniDocBench 1.6~\cite{omnidocbench}, as shown in Tab.~\ref{tab:omni16_full_performance}. MonkeyOCRv2-Parsing surpasses much larger general-purpose VLMs, including Qwen3-VL-235B, GPT-5.2, InternVL3.5-241B, and Kimi K2.5, but still lags behind
the
latest specialized document parsing models including GLM-OCR, 
MinerU2.5-pro, PaddleOCR-VL-1.6, and HunyuanOCR-1.5.

Tab.~\ref{tab:omni16_full_performance} is a system-level comparison. The compared systems differ in training data, post-training, layout modules, region-refinement strategies, and inference pipelines; therefore, this comparison cannot identify the cause of the remaining performance gap or isolate the contribution of the visual encoder. We use the controlled comparison in Tab.~\ref{tab:doc_understanding_results}, rather than Tab.~\ref{tab:omni16_full_performance}, for encoder-level attribution.

\subsection{Document Understanding}
\label{sec:doc_understanding}
    Document understanding aims to comprehend and answer questions about the visual content of scanned documents, tables, and charts. To isolate the effect of the visual backbone, we build VLMs that pair different vision encoders with the same Qwen3-1.7B~\cite{qwen3} language model through an MLP projector, following the conventional VLM framework~\cite{llava1.5}.

    \textbf{Training Details.} Training proceeds in two stages. We first only train the MLP projector with a learning rate of $1 \times 10^{-3}$, then jointly optimize the projector and the language model with a learning rate of $1 \times 10^{-5}$. Both stages use a batch size of $128$, and the vision encoder is kept frozen throughout.
    
    \textbf{Evaluation.} Following the standard setting~\cite{textmonkey}, we evaluate our models on eight representative document VQA benchmarks: 
    DocVQA~\cite{docvqa}, InfoVQA~\cite{infovqa}, DeepForm (DF)~\cite{deepform}, KLC~\cite{klc}, WTQ~\cite{wtq}, ChartQA~\cite{chartqa}, DT-VQA~\cite{dtvqa} and OCRBench~\cite{ocrbench}. As shown in Tab.~\ref{tab:doc_understanding_results}, all backbones are paired with the same Qwen3-1.7B LLM and trained on identical data, optimization schedules, and decoding procedures, while each encoder uses its native input configuration reported in App.~\ref{sec:appendix_vqa_config}.
    Under this controlled setting, the VLM equipped with MonkeyOCRv2-B consistently achieves stronger document-understanding performance than the other pretrained encoders, reaching an average score of 57.2 across the eight benchmarks. The strongest prior vision encoders in this comparison are OpenVision-B (44.0) and RADIOv2.5-B (37.5), both of which adopt generative or multi-teacher training paradigms that extend beyond pure global semantic alignment; nevertheless, MonkeyOCRv2 surpasses them by 13.2 and 19.7 absolute, respectively. Encoders primarily optimized for global semantics or region-level segmentation perform substantially worse, including SAM (25.2), SigLIP 2 (24.9), and DINOv3 (16.1). 
    Image reconstruction encourages the vision encoder to preserve additional fine-grained visual information that may be discarded under text-only pretraining.
    Incorporating the MSE reconstruction loss improves performance by 1.0\%. Furthermore, introducing edge- and distance-aware reconstruction losses enables better modeling of text strokes, yielding an additional 4.2\% improvement.
    Overall, these results demonstrate that MonkeyOCRv2 produces significantly stronger visual representations for document images, enabling more effective visual encoding of structured text, layouts, and semantic cues in complex document understanding scenarios.


\begin{table*}[t]
\centering
\setlength{\tabcolsep}{4pt}
\caption{
Controlled comparison of vision foundation models on document understanding. 
All downstream components, training data, optimization settings, and decoding procedures are held fixed. The vision encoder is the only component intentionally varied; exact input resolutions, patch sizes, and visual-token counts are reported in App.~\ref{sec:appendix_vqa_config}.
The proposed MonkeyOCRv2 achieves the best results across all benchmarks. oCLIP, DiT and ours are pretrained on document images. Baseline denotes pretraining without the image reconstruction objective. * indicates the use of edge- and distance-aware reconstruction losses.
}
\label{tab:doc_understanding_results}
\resizebox{1\textwidth}{!}{
\begin{tabular}{l|c|ccccccccc}
\toprule
\textbf{Vision Encoder} &
\textbf{Params} &
\textbf{Overall} &
\textbf{DocVQA~\cite{docvqa}} &
\textbf{InfoVQA~\cite{infovqa}} &
\textbf{DF~\cite{deepform}} &
\textbf{KLC~\cite{klc}} &
\textbf{WTQ~\cite{wtq}} &
\textbf{ChartQA~\cite{chartqa}} &
\textbf{DT-VQA~\cite{dtvqa}} &
\textbf{OCRBench~\cite{ocrbench}} \\
\midrule
\multicolumn{11}{l}{\textit{Frozen vision encoders paired with the same Qwen3-1.7B (controlled comparison)}} \\
\midrule
CLIP-B~\cite{clip} & 86M & 16.0 & 20.1 & 24.2 & 2.3 & 13.8 & 12.8 & 22.2 & 22.3 & 10.6 \\
SigLIP 2-B~\cite{siglip2} & 93M & 24.9 & 27.0 & 23.5 & 3.1 & 16.7 & 17.4 & 35.0 & 41.5 & 35.1 \\
RADIOv2.5-B~\cite{radio2d5} & 98M & 37.5 & 60.3 & 31.2 & 29.9 & 30.4 & 29.7 & 51.1 & 44.2 & 23.1 \\
OpenVision-B~\cite{openvision} & 87M & 44.0 & 63.3 & 30.7 & 19.8 & 33.1 & 31.1 & {58.3} & {62.6} & \underline{52.9} \\
DINOv3-B~\cite{dinov3} & 86M & 16.1 & 26.5 & 20.8 & 5.6 & 13.2 & 14.0 & 28.9 & 15.8 & 3.9 \\
SAM-B~\cite{sam} & 90M & 25.2 & 37.8 & 22.2 & 4.7 & 17.5 & 17.6 & 46.5 & 33.3 & 21.9 \\
SAM2-B~\cite{sam2} & 69M & 22.3 & 32.5 & 21.9 & 2.7 & 15.8 & 16.6 & 40.2 & 30.3 & 18.4 \\
oCLIP~\cite{oclip} & 24M & 12.4 & 14.8 & 19.5 & 1.4 & 7.4 & 11.4 & 17.9 & 19.2 & 7.4 \\
DiT~\cite{dit} & 86M & 8.9 & 11.3 & 20.9 & 0.9 & 5.2 & 9.9 & 12.0 & 9.2 & 1.9 \\
\midrule
MonkeyOCRv2-S (Baseline) & 28M &{50.7} &{70.5} & {37.0} & {60.6} & {35.2} & {36.7} & {57.4} & {57.3} & {50.7} \\
MonkeyOCRv2-S (MSE only) & 28M &{51.7} &{71.0} & {37.5} & {62.7} & {35.8} & {38.7} & {58.3} & {58.7} & {50.9} \\
MonkeyOCRv2-S* & 28M & \underline{55.9} & \textbf{79.3} & \underline{44.5} & \underline{65.1} & \underline{37.6} & \underline{43.0} & \textbf{62.0} & \underline{63.1} & {52.2} \\ \midrule
MonkeyOCRv2-B* & 113M & \textbf{57.2} & \textbf{79.3} & \textbf{46.3} & \textbf{65.8} & \textbf{38.2} & \textbf{43.2} & \textbf{62.0} & \textbf{64.3} & \textbf{58.1} \\
\bottomrule

\end{tabular}}
\end{table*}

\section{Discussion}

\begin{figure*}[t]
    \centering
    \includegraphics[width=\textwidth]{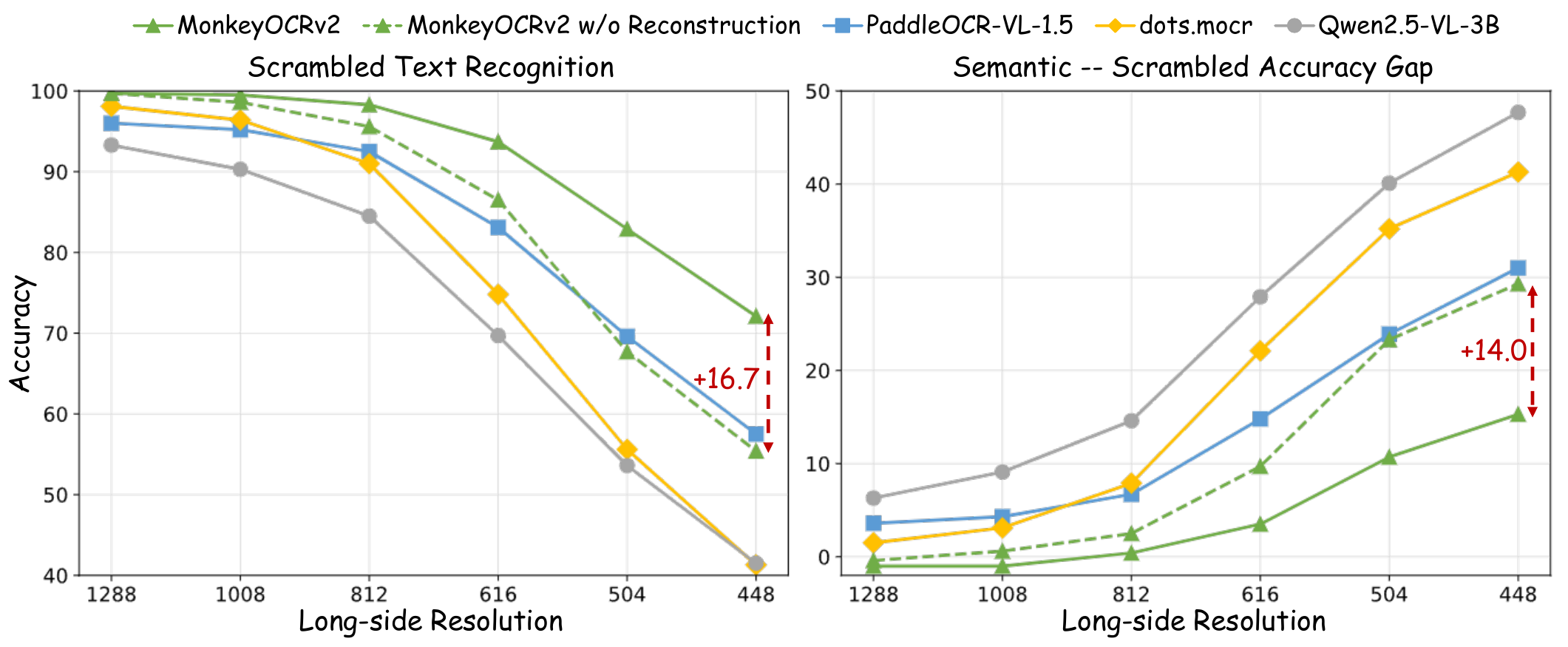}
    \caption{Left: scrambled text recognition accuracy. Right: the accuracy gap between semantically coherent and scrambled text, where a smaller gap indicates weaker dependence on linguistic context under this controlled perturbation. Image reconstruction improves scrambled text recognition and narrows the semantic--scrambled accuracy gap, especially at low resolution.
    }
\label{fig:resolution_comparison}
\end{figure*}

\subsection{Analysis of Dependence on Linguistic Context}
\label{subsec:alpd}
    High recognition accuracy on semantically coherent text does not necessarily imply strong visual perception, because a decoder can exploit linguistic context to recover visually ambiguous characters. To examine this effect, we construct two paired evaluation settings: semantically coherent text and randomly scrambled text. The former preserves natural linguistic context, whereas the latter removes most of it by shuffling characters, requiring the model to rely more heavily on character-level visual evidence. We progressively reduce the input resolution from 1288 to 448 and report two metrics in Fig.~\ref{fig:resolution_comparison}: scrambled-text recognition accuracy and the accuracy gap between semantically coherent and scrambled text. We use this gap as an operational proxy for dependence on linguistic context. Because character scrambling also introduces an out-of-distribution input distribution and may interact with tokenization and decoding, the gap should not be interpreted as a complete measure of hallucination or language-prior dependence.

    The results show that existing VLMs become substantially less robust when linguistic context is removed, especially as the input resolution decreases. As the resolution drops from 1288 to 448, scrambled-text accuracy decreases by 38.5\% absolute for PaddleOCR-VL-1.5, 56.8\% for dots.mocr, and 51.8\% for Qwen2.5-VL-3B. The semantic--scrambled accuracy gap also widens as the visual input is degraded. For example, the gap for PaddleOCR-VL-1.5 increases from 3.6\% at full resolution to 31.0\% at a resolution of 448. Under this controlled perturbation, the widening gap suggests that the model depends more heavily on linguistic context when the visual signal becomes weak.
    
    MonkeyOCRv2-S-Parsing is more robust to resolution degradation, and the pixel-level reconstruction objective provides additional gains when linguistic context is removed. Without reconstruction, scrambled-text accuracy decreases from 99.7\% to 55.4\% as the resolution drops from 1288 to 448, corresponding to a reduction of 44.3\% absolute. With reconstruction, the accuracy remains at 72.1\% at the lowest resolution, reducing the drop to 27.6\%. Correspondingly, the semantic--scrambled accuracy gap at a resolution of 448 decreases from 29.3\% without reconstruction to 15.3\% with reconstruction. These results are consistent with reconstruction preserving fine-grained visual details such as strokes, contours, and character structures. The benefit is smaller but remains consistent on semantically coherent text: at resolutions of 616, 504, and 448, reconstruction improves accuracy from 96.2\% to 97.2\%, 91.0\% to 93.6\%, and 84.7\% to 87.4\%, respectively.

    Overall, these results provide evidence that pixel-level reconstruction improves robustness when linguistic context is unavailable by encouraging the encoder to preserve fine-grained visual information. Importantly, the scrambled-text metrics are measured on the full parsing pipeline rather than on the encoder in isolation. The smaller accuracy drop and narrower semantic--scrambled gap therefore indicate that the visual information retained by the encoder remains useful after language decoding. Nevertheless, these metrics are operational proxies rather than complete measures of hallucination, because character scrambling introduces distribution shift and may interact with tokenization and decoding.

\subsection{Document Hallucination Evaluation on CHAOS-Bench}

CHAOS-Bench~\cite{hunyuanocr15} is a benchmark for evaluating document hallucination and output faithfulness. It constructs test samples by modifying one character in 2–3 selected words on each document page, turning them into visually observable but semantically meaningless words. Models are then required to parse the modified document images, and the benchmark measures whether these perturbed words are faithfully preserved in the output. The evaluation metric is the page-average recall of the perturbed words, which reflects the model's ability to rely on visual evidence rather than language priors when the two are in conflict.

As shown in Tab.~\ref{tab:chaos_bench_results}, MonkeyOCRv2-B-Parsing achieves the best performance, outperforming HunyuanOCR-1.5 by 3.7\% and PaddleOCR-VL-1.6 by 11.9\%. These results indicate that, compared with existing models, MonkeyOCRv2-B-Parsing relies more on visual evidence than language priors when recognizing text. Moreover, MonkeyOCRv2-S-Parsing also surpasses HunyuanOCR-1.5 by 0.5\%. Compared with the baseline that uses a MonkeyOCRv2-S encoder pretrained without the image reconstruction objective, incorporating image reconstruction improves performance by 2.6\%. This demonstrates that the image reconstruction objective helps preserve fine-grained visual evidence and mitigates the tendency of the visual encoder to learn semantic shortcuts during pretraining.

\begin{table}[t]
    \centering
    \small
    \caption{
    \textbf{Results on CHAOS-Bench~\cite{hunyuanocr15}.}
    We report the page-average recall of perturbed words, which measures output faithfulness when visual evidence conflicts with language priors. Higher is better.
    \emph{Baseline} denotes the model's visual encoder is pretrained without image reconstruction. All other results are taken from CHAOS-Bench.
    }
    \label{tab:chaos_bench_results}
    \setlength{\tabcolsep}{8pt}
    \begin{tabular}{lcc}
        \toprule
        \textbf{Model} &
        \textbf{Size} &
        \textbf{Page-average Recall $\uparrow$} \\
        \midrule
        dots.ocr~\cite{dotsocr} & 3B   & 3.0 \\
        GLM-OCR~\cite{glm-ocr} & 1B   & 5.8 \\
        PaddleOCR-VL-1.6~\cite{ppocrvl16} & 0.9B & 6.0 \\
        DeepSeek-OCR 2~\cite{deepseekocr2} & 3B   & 6.3 \\
        MinerU2.5Pro~\cite{mineru2.5pro} & 1.2B & 6.3 \\
        HunyuanOCR-1.5~\cite{hunyuanocr15} & 1B & 14.2\\
        \midrule
        MonkeyOCRv2-S-Parsing (\emph{Baseline}) & 0.6B & 12.1\\
        MonkeyOCRv2-S-Parsing  & 0.6B & 14.7 (\textcolor{red}{+2.6})\\ \midrule
        MonkeyOCRv2-B-Parsing  & 0.7B & \textbf{17.9}\\

        \bottomrule
    \end{tabular}
\end{table}

\begin{figure*}[t]
    \centering
    \includegraphics[width=\linewidth]{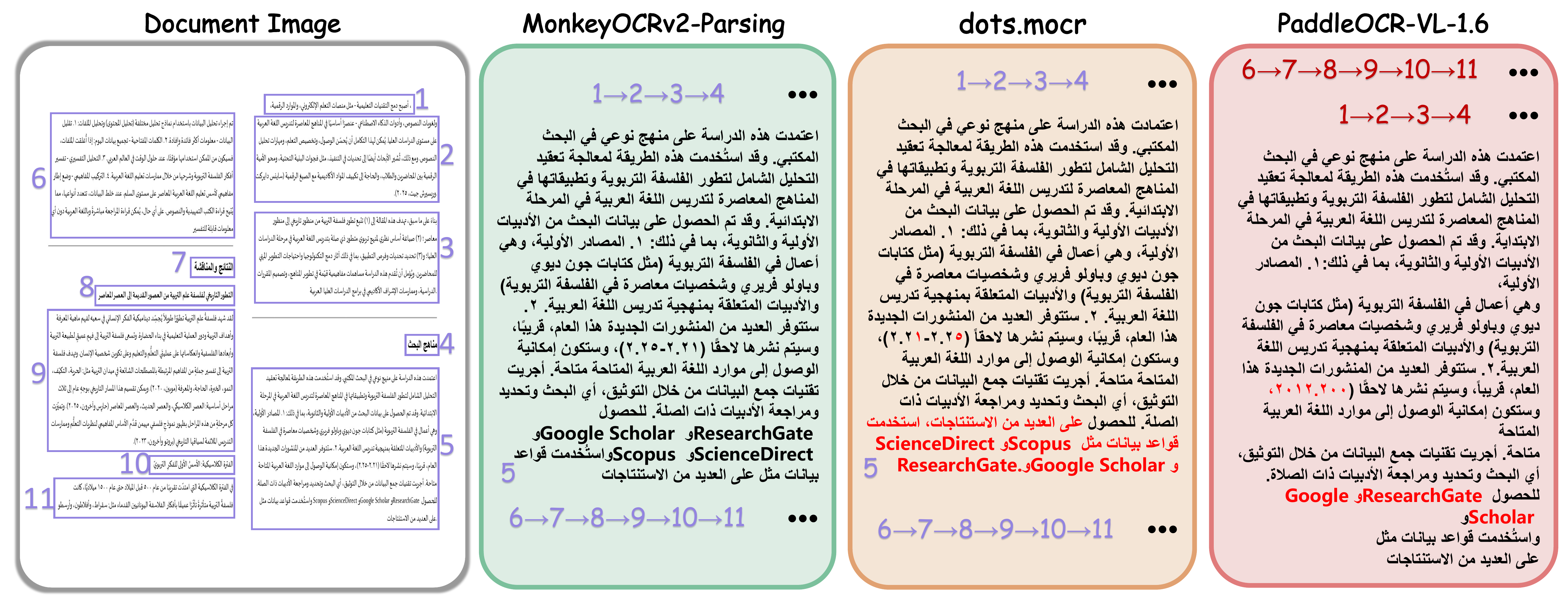}
    
    \caption{Visualization comparisons with leading document parsing models on an Arabic document that requires right-to-left reading. Incorrect predictions are highlighted in red.}
    \label{fig:parse_ar}
\end{figure*}

\begin{figure*}[t]
    \centering
    \includegraphics[width=\linewidth]{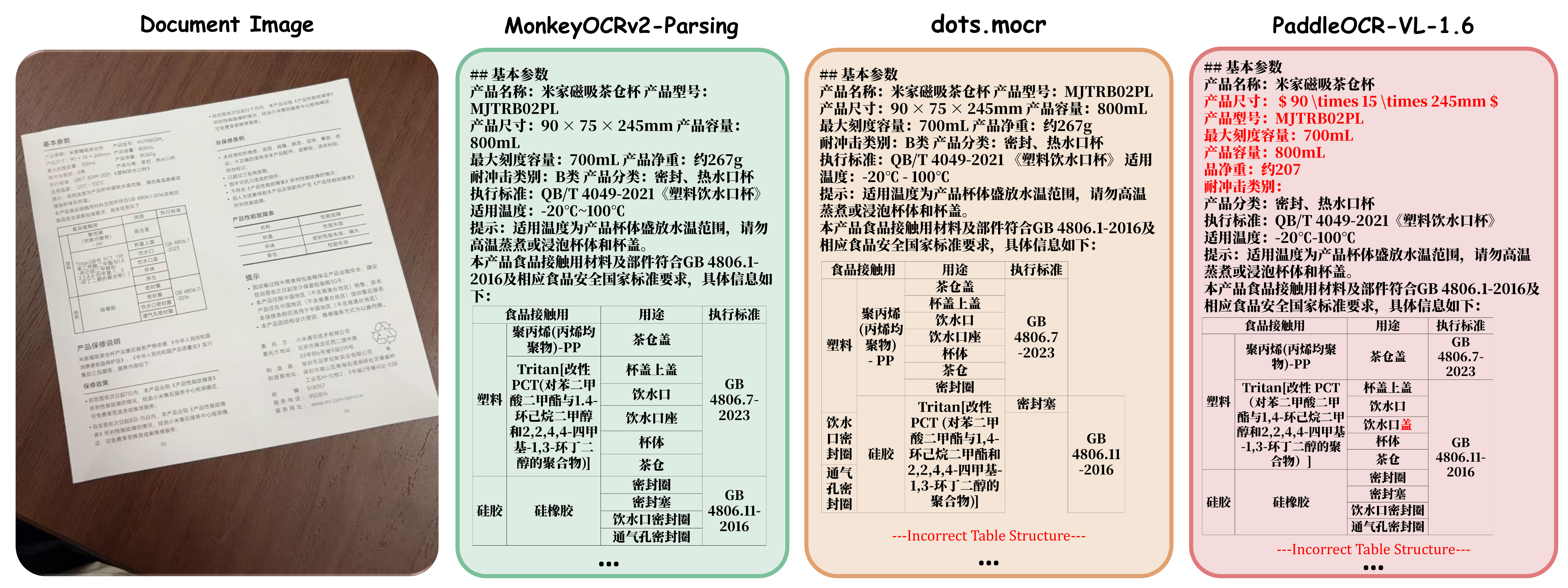}
    
    \caption{Visualization comparisons with leading document parsing models on a photographed Chinese instruction manual.}
    \label{fig:parse_zh}
\end{figure*}

\subsection{Qualitative Comparison on Document Parsing}

As shown in Fig.~\ref{fig:parse_ar} and Fig.~\ref{fig:parse_zh}, we compare MonkeyOCRv2-Parsing with other leading document parsing models. For the two-column Arabic document in Fig.~\ref{fig:parse_ar}, PaddleOCR-VL-1.6 fails to follow the right-to-left reading order of the Arabic layout and even generates text in irrelevant languages. Although dots.mocr captures the correct reading order, it still suffers from misordered text and hallucinated outputs in mixed Arabic--English scenarios. For the photographed Chinese document in Fig.~\ref{fig:parse_zh}, PaddleOCR-VL-1.6 misses portions of the content and fails to reconstruct the table structure accurately; dots.mocr likewise produces incorrect table structures. In contrast, MonkeyOCRv2-Parsing delivers more accurate parsing results on both documents.

\begin{figure*}[t]
    \centering
    \includegraphics[width=\textwidth]{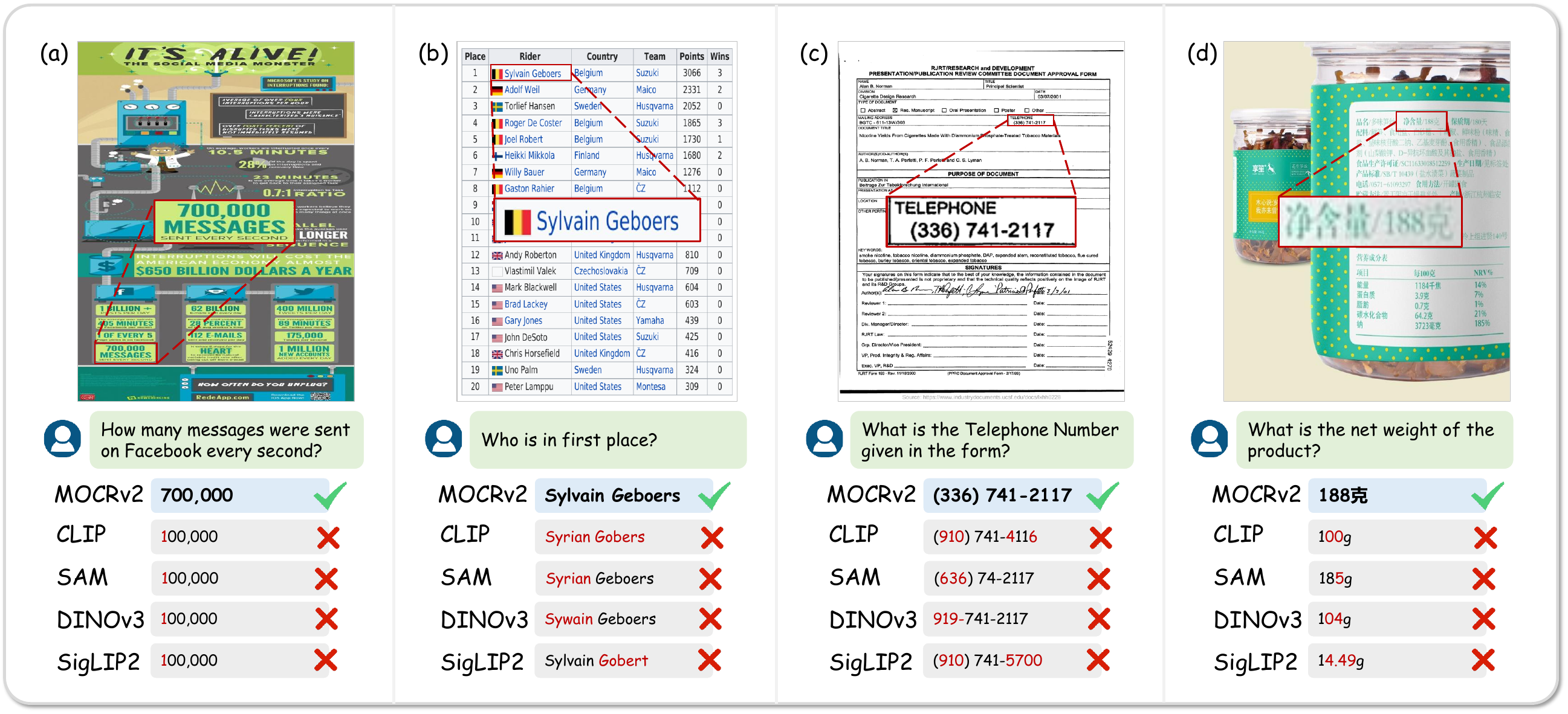}
    \caption{Visualization comparisons with popular vision foundation models on document understanding tasks. MonkeyOCRv2 demonstrates stronger fine-grained text perception capabilities. MOCRv2: MonkeyOCRv2.}
\label{fig:vqa}
\end{figure*}

\subsection{Qualitative Comparison on Document Understanding}

As illustrated in Fig.~\ref{fig:vqa}, we present qualitative comparisons of different vision foundation models on document understanding tasks. MonkeyOCRv2 demonstrates substantially stronger fine-grained text perception, enabling accurate recognition of dense textual content, including numerical statistics in infographics, rider names in forms, telephone numbers in documents, and Chinese text on product labels. In contrast, CLIP and other vision foundation models frequently confuse visually similar characters and struggle to recognize complete text sequences. For example, CLIP incorrectly recognizes 
``700,000'' as ``100,000'', 
misreads the rider name ``Sylvain Geboers'' as ``Syrian Gobers'', fails to accurately identify telephone numbers and product net weight, and produces character-level recognition errors. Although some models can roughly localize relevant text regions, they still struggle with precise text recognition. These observations suggest that general-purpose vision foundation models primarily learn representations optimized for global semantic understanding, whereas MonkeyOCRv2 learns richer fine-grained visual representations that preserve character-level details essential for document understanding.

\section{Limitations}
Our study has several limitations. First, MonkeyOCRv2-Parsing adopts a deliberately minimal supervised fine-tuning pipeline with the encoder frozen, without the progressive post-training used by leading specialized parsers; on saturated benchmarks such as OmniDocBench this leaves a gap to the strongest task-specific systems, even though the same encoder reaches open-source state-of-the-art on the more discriminative MDPBench. Second, our parsing architecture predicts layout autoregressively and recognizes each element in a separate pass, which favors accuracy and structural fidelity over inference speed; latency-oriented deployment would benefit from a more parallel layout stage. Third, the present work establishes that a document-oriented, reconstruction-aware objective improves fine-grained perception, but does not yet isolate the individual contribution of decoder design or the reconstruction-weight schedule; a systematic study of these factors is left for future work. Finally, although MonkeyDoc v2 spans 17 languages, its coverage remains skewed toward high-resource scripts, and extending balanced supervision to low-resource and historical writing systems is an important direction.

\section{Conclusion}
In this paper, we introduced MonkeyOCRv2, a visual foundation model for document intelligence. Unlike general-purpose encoders pretrained on natural images, MonkeyOCRv2 is built to preserve the fine-grained textual detail and layout structure of document images. 
By jointly optimizing autoregressive text generation and pixel-level image reconstruction, it learns representations that capture textual semantics while retaining character strokes, glyph shapes, and local visual details, thereby improving robustness when linguistic context is weak or unavailable.
To supply the supervision such training demands, we constructed MonkeyDoc v2, a large-scale multilingual document corpus of 113 million samples across 17 languages. 
Whether integrated into existing systems as a backbone substitution or paired, frozen, with lightweight language models, MonkeyOCRv2 yields consistent gains across seven tasks:
multilingual document parsing, document understanding, text recognition, formula recognition, document tampering detection, scene text detection, and overlapping text segmentation.

Beyond these specific gains, our results carry a broader message. Document intelligence has long relied on encoders built for object and scene recognition, lacking one designed for its own visual statistics. 
MonkeyOCRv2 shows that a single document-oriented pre-training recipe, instantiated as a compact encoder family, transfers across seven disparate systems and improves each, 
and that a pixel-level reconstruction objective improves recognition when linguistic context is removed and narrows the semantic--scrambled accuracy gap.
This suggests that document-oriented visual pre-training can serve as a foundation for document intelligence in its own right, rather than a domain to be served by encoders built for natural scenes.
We hope MonkeyOCRv2 and MonkeyDoc v2 help establish text as a first-class visual modality and provide reusable foundations for multilingual OCR, document understanding, and broader document-intelligence systems.

\thispagestyle{plain}
\bibliographystyle{plain}
\bibliography{monkeyocr}

\clearpage
\appendix
\section{Data Distribution of MonkeyDoc v2}
\label{sec:appendix_data}

\begin{figure*}[h]
    \centering
    \includegraphics[width=\linewidth]{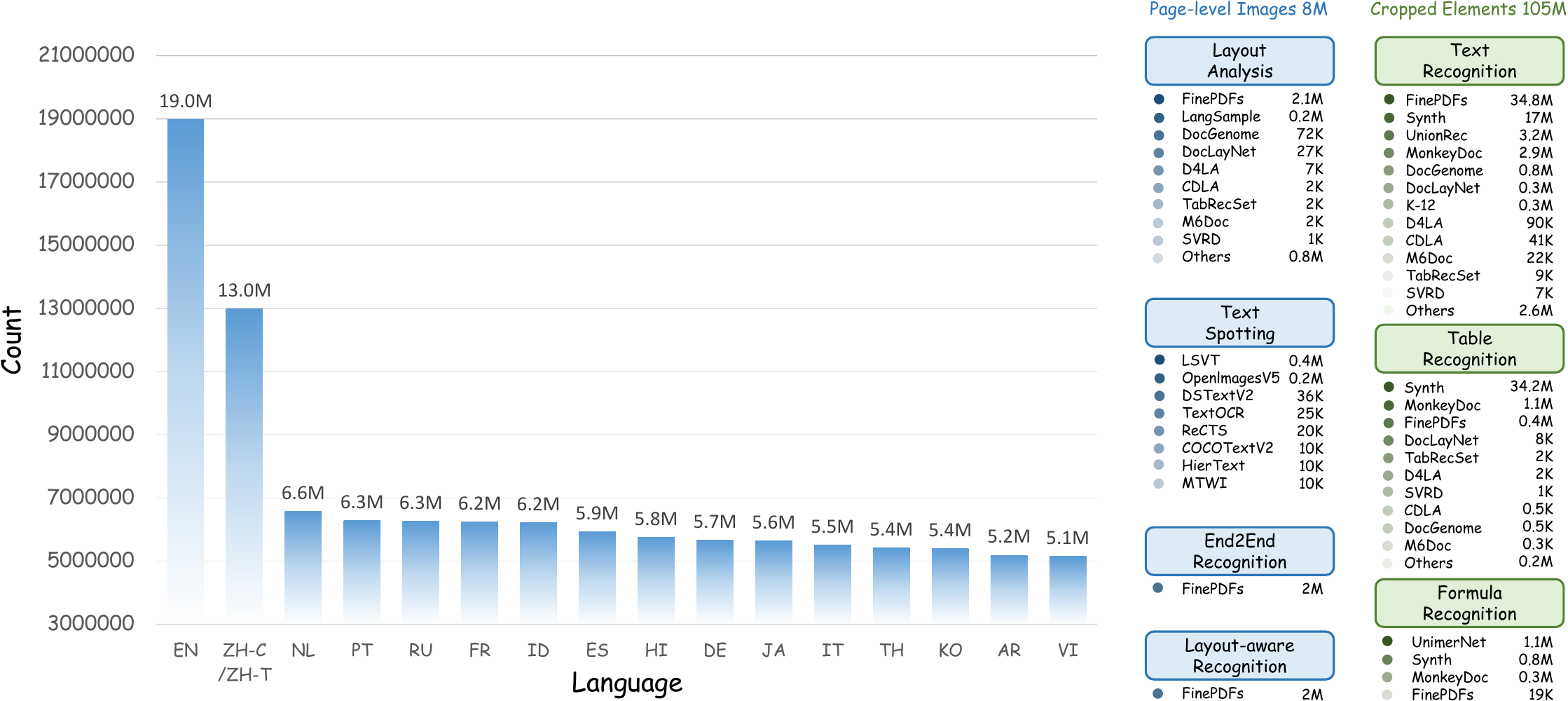}
    \caption{Detailed data distribution of MonkeyDoc v2.}
    \label{fig:data_distribution}
\end{figure*}

Fig.~\ref{fig:data_distribution} provides a detailed breakdown of MonkeyDoc v2.
At the language level, the corpus covers 17 languages and is dominated by English and Chinese, with 19M English samples and 13M Simplified/Traditional Chinese samples, followed by Dutch, Portuguese, Russian, French, Indonesian, Spanish, Hindi, German, Japanese, Italian, Thai, Korean, Arabic, and Vietnamese. The data are organized into two granularities: 8M page-level images for page-level
images and 105M cropped elements for fine-grained recognition tasks. The page-level subset mainly supports layout analysis, end-to-end recognition,
layout-aware recognition and text spotting, while the cropped-element subset supports text recognition, table recognition, and formula recognition.

\section{Integration Protocol for Downstream Tasks}
\label{sec:appendix_integration}
\textbf{Text Recognition.} We replace the original visual encoders in both CRNN~\cite{crnn} and PARSeq~\cite{parseq} with the MonkeyOCRv2-S visual backbone. For PARSeq, the MonkeyOCRv2-S encoder directly produces 384-dimensional visual tokens, which are used as the memory tokens for the Transformer decoder cross-attention, while the original PARSeq decoder, autoregressive prediction, and iterative refinement strategy are kept unchanged. For CRNN, we convert the two-dimensional MonkeyOCRv2-S visual token map into a one-dimensional sequence by applying a learnable height-wise pooling layer, and then feed the resulting width-wise feature sequence into the original BiLSTM~\cite{bilstm} recognition head with CTC~\cite{ctc} supervision. 
Following the SVTRv2~\cite{svtrv2} training protocol, we use a learning rate of $7\times10^{-4}$ and keep the total number of training epochs unchanged. Training is performed in two stages: we first freeze the visual encoder and train only the recognition head to preserve the pretrained visual representations, and then jointly fine-tune the entire model. The two stages last for 20/20 epochs on English datasets and 30/70 epochs on Chinese datasets, respectively.

\textbf{Formula Recognition.} 
We replace the original UniMERNet-T~\cite{unimernet} visual encoder with the MonkeyOCRv2-S backbone, while keeping the original MBart decoder unchanged. Since our encoder outputs 384-dimensional visual tokens whereas the MBart~\cite{mbart} decoder uses 512-dimensional hidden
states, we insert a learnable linear encoder-to-decoder projection layer to map the visual features from 384 to 512 dimensions before decoder cross-attention. 
Apart from the required 384-to-512 interface projection and the encoder warm-up schedule, the decoder architecture, training data, and optimization configuration are identical to the controlled baseline.

\textbf{Text Detection.} To integrate the MonkeyOCRv2-AS backbone into downstream scene text detectors, stage-wise token embeddings are reshaped into four stages of 2D feature maps with strides $4/8/16/32$ and channels $[64, 128, 256, 512]$, respectively. For DBNet~\cite{dbnet} and PSENet~\cite{wang2019shape}, all four stages are used to feed FPN necks. For DPText-DETR~\cite{dptextdetr}, the last three stages feed a 6-layer transformer encoder. All detectors are fine-tuned end-to-end using the AdamW optimizer with a learning rate of $10^{-4}$ and a weight decay of $10^{-4}$.

\textbf{Document Tampering Detection.} Starting from the official FFDN~\cite{chen2024enhancing} implementation, we replace its ConvNeXt-V2-Base~\cite{convnextv2} backbone with either (i) a ViTAEv2-S checkpoint pretrained by DeepSolo or (ii) MonkeyOCRv2-AS. For both variants, the first two stages form the Visual Perception Head and extract tampering cues from RGB features, while the two deeper stages encode frequency-enhanced features, yielding a four-scale feature pyramid. The learning rate of the Visual Perception Head is set to 0.2$\times$ the base learning rate. All other data, augmentations, optimization settings, and evaluation procedures are kept identical.

\textbf{Overlapping Text Segmentation.} For Mask2Former, the initial learning rate and batch size are set to $1\times 10^{-4}$ and 8, respectively, while they are configured as $2\times 10^{-4}$ and 4 for MOTS. Both models are optimized using the AdamW optimizer with a weight decay of 0.05 on $512\times 512$ cropped inputs, following a poly learning rate policy.

\section{Text Recognition on Common Benchmarks}
\label{sec:appendix_common}

\begin{table*}[h]
\centering
\setlength{\tabcolsep}{4pt}
\caption{Integrating MonkeyOCRv2-S into the representative CRNN and the leading PARSeq models improves performance on common benchmarks.}
\label{tab:common_results}
\resizebox{0.7\textwidth}{!}{
\begin{tabular}{l c cccccc}
\toprule
Model & Avg & IIIT5k~\cite{IIIT5K} & SVT~\cite{svt} & IC13~\cite{ic13} & IC15~\cite{ic15}
&
SVTP~\cite{SVTP}
&
CUTE80~\cite{cute80}
\\
\midrule

ABINet~\cite{abinet}
& 95.8
& 98.5
& 98.1
& 97.7
& 90.1
& 94.1
& 96.5
\\

MAERec~\cite{union14m}
& 96.4
& \textbf{99.2}
& 97.8
& 98.2
& 90.4
& 94.3
& 98.3
\\

CPPD~\cite{cpdd}
& 96.4
& 99.0
& 97.8
& 98.2
& 90.4
& 94.0
& \textbf{99.0}
\\

IGTR-AR~\cite{igtrar}
& 96.5
& 98.7
& \textbf{98.4}
& 98.1
& 90.5
& 94.9
& 98.3
\\

SMTR~\cite{smtr}
& 95.9
& 99.0
& 97.4
& 98.3
& 90.1
& 92.7
& 97.9
\\

SVTRv2~\cite{svtrv2}
& 96.6
& \textbf{99.2}
& 98.0
& \textbf{98.7}
& \textbf{91.1}
& 93.5
& \textbf{99.0}
\\

\midrule

CRNN~\cite{crnn} (ResNet)
& 90.2
& 95.8
& 91.8
& 94.6
& 84.9
& 83.1
& 91.0
\\

CRNN (MonkeyOCRv2-S)
& 92.5
& 97.4
& 94.3
& 96.5
& 86.8
& 87.1
& 93.1
\\

\midrule

PARSeq~\cite{parseq} (ViT)
& 96.4
& 98.9
& 98.1
& 98.4
& 90.1
& 94.3
& 98.6
\\

PARSeq (MonkeyOCRv2-S)
& \textbf{96.8}
& \textbf{99.2}
& 98.0
& 98.5
& 90.4
& \textbf{96.1}
& 98.6
\\

\bottomrule
\end{tabular}
}
\end{table*}
We also evaluate on six widely used regular and irregular scene text benchmarks (Common), including ICDAR2013~\cite{ic13} (IC13), SVT~\cite{svt}, IIIT5K~\cite{IIIT5K}, ICDAR2015~\cite{ic15} (IC15), SVTP~\cite{SVTP}, and CUTE80~\cite{cute80}. As these benchmarks are already close to saturation, only modest improvements are expected. As shown in Tab.~\ref{tab:common_results}, replacing the original visual encoder of CRNN with MonkeyOCRv2-S consistently improves performance across all Common benchmarks, yielding an average gain of 2.3\%. Despite the near-saturated performance of PARSeq, integrating MonkeyOCRv2-S still brings a further average improvement of 0.4\%, surpassing the previous state-of-the-art, SVTRv2.

\section{Vision Encoder Configuration on Document Understanding}
\label{sec:appendix_vqa_config}

Tab.~\ref{tab:vqa_config} summarizes the input resolution settings, patch sizes of different vision encoders and the average number of visual tokens produced by each encoder over all images from the 8 evaluated benchmarks in Sec.~\ref{sec:doc_understanding}. All encoders are evaluated using the input resolution settings supported during their original training. For example, SAM is trained at a fixed resolution of 1024, whereas SigLIP 2 (naflex) preserves the original image resolution and aspect ratio. 

\begin{table*}[h]
\centering
\setlength{\tabcolsep}{4pt}
\caption{Configuration of different vision encoders and average visual tokens on 8 benchmarks used for evaluation in Sec.~\ref{sec:doc_understanding}. Any denotes using the original image resolution as input. For fixed-resolution settings, both the height and width are resized to the specified resolution. All encoders follow their original training input configurations.}

\label{tab:vqa_config}
\resizebox{0.5\textwidth}{!}{
\begin{tabular}{l c cccccc}
\toprule
Vision encoder &  Resolution  &  Patch size  &  Avg. visual tokens
\\
\midrule
CLIP~\cite{clip}
& 224
& 16
& 196
\\
Siglip 2~\cite{siglip2}
& Any
& 16
& 825
\\
RADIOv2.5~\cite{radio2d5}
& Any
& 16
& 825
\\
OpenVision~\cite{openvision}
& 384
& 8
& 2304
\\
DINOv3~\cite{dinov3}
& Any
& 16
& 825
\\
SAM~\cite{sam}
& 1024
& 16
& 4096
\\
SAM2~\cite{sam2}
& 1024
& 16
& 4096
\\
DiT~\cite{dit}
& 224
& 16
& 196
\\
oCLIP~\cite{oclip}
& 512
& 16
& 1024
\\
\midrule
MonkeyOCRv2-B
& Any
& 14
& 1082
\\
MonkeyOCRv2-S
& Any
& 14
& 1082
\\
\bottomrule
\end{tabular}
}
\end{table*}

\section{Examples from MonkeyDoc v2}
\label{app:training-data-examples}

\begin{figure}[H]
    \centering
    \includegraphics[width=\textwidth,height=0.9\textheight,keepaspectratio]{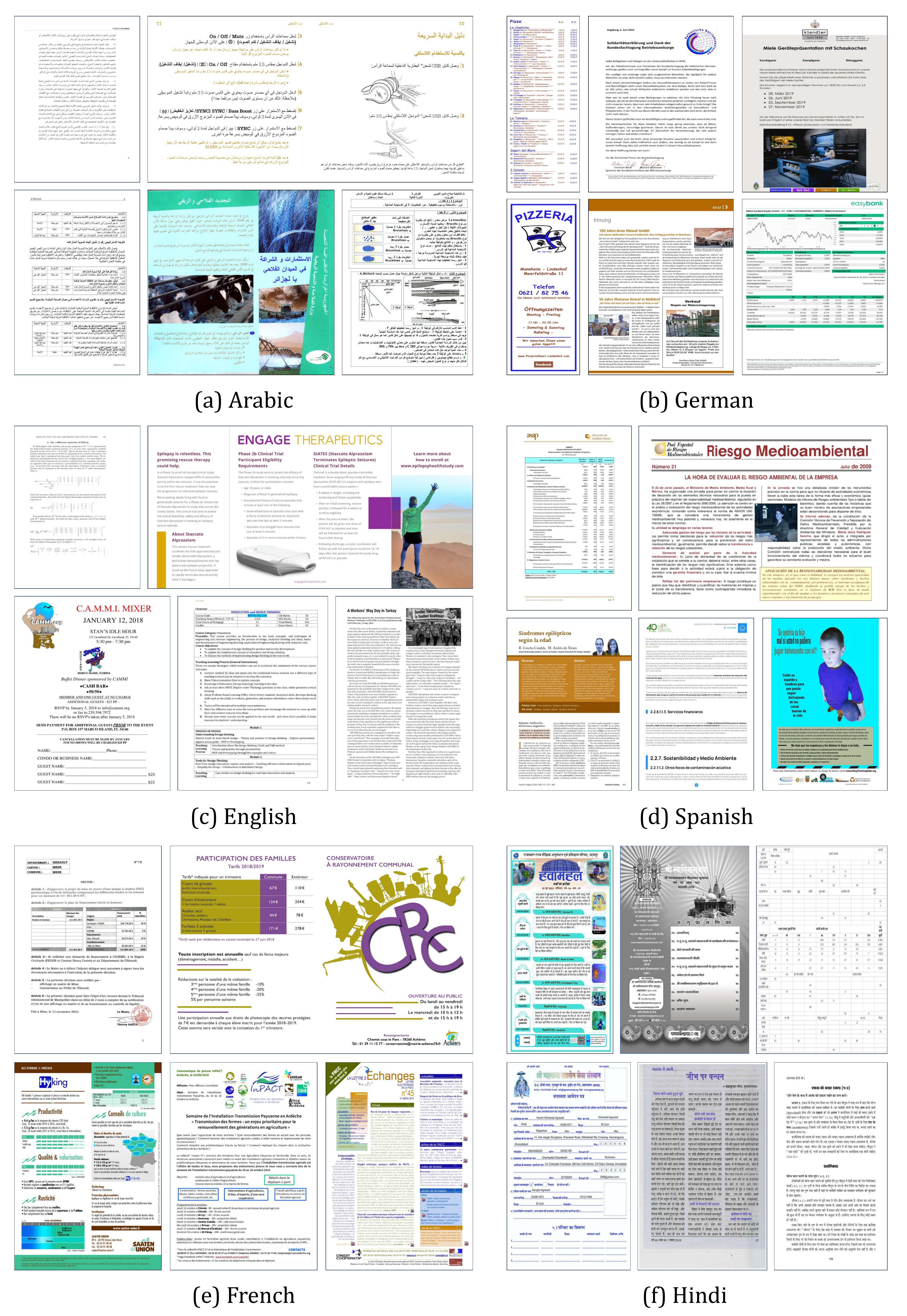}
    \caption{Visualization of Arabic, German, English, Spanish, French, and Hindi images in MonkeyDoc v2.}
    \label{fig:app-e-training-examples-group1}
\end{figure}

\begin{figure}[p]
    \centering
    \includegraphics[width=\textwidth,height=\textheight,keepaspectratio]{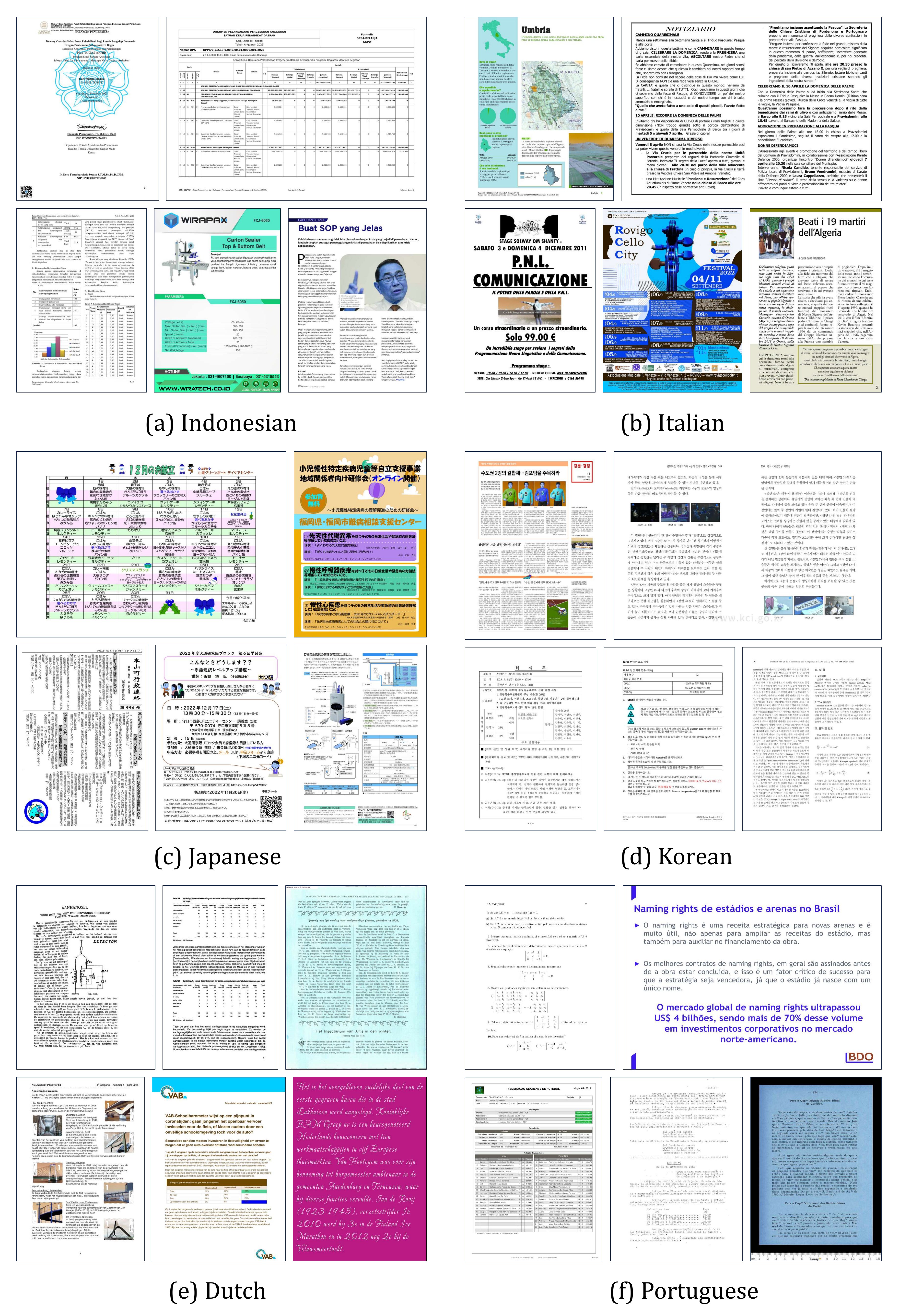}
    \caption{Visualization of Indonesian, Italian, Japanese, Korean, Dutch, and Portuguese images in MonkeyDoc v2.}
    \label{fig:app-e-training-examples-group2}
\end{figure}

\begin{figure}[p]
    \centering
    \includegraphics[width=\textwidth,height=\textheight,keepaspectratio]{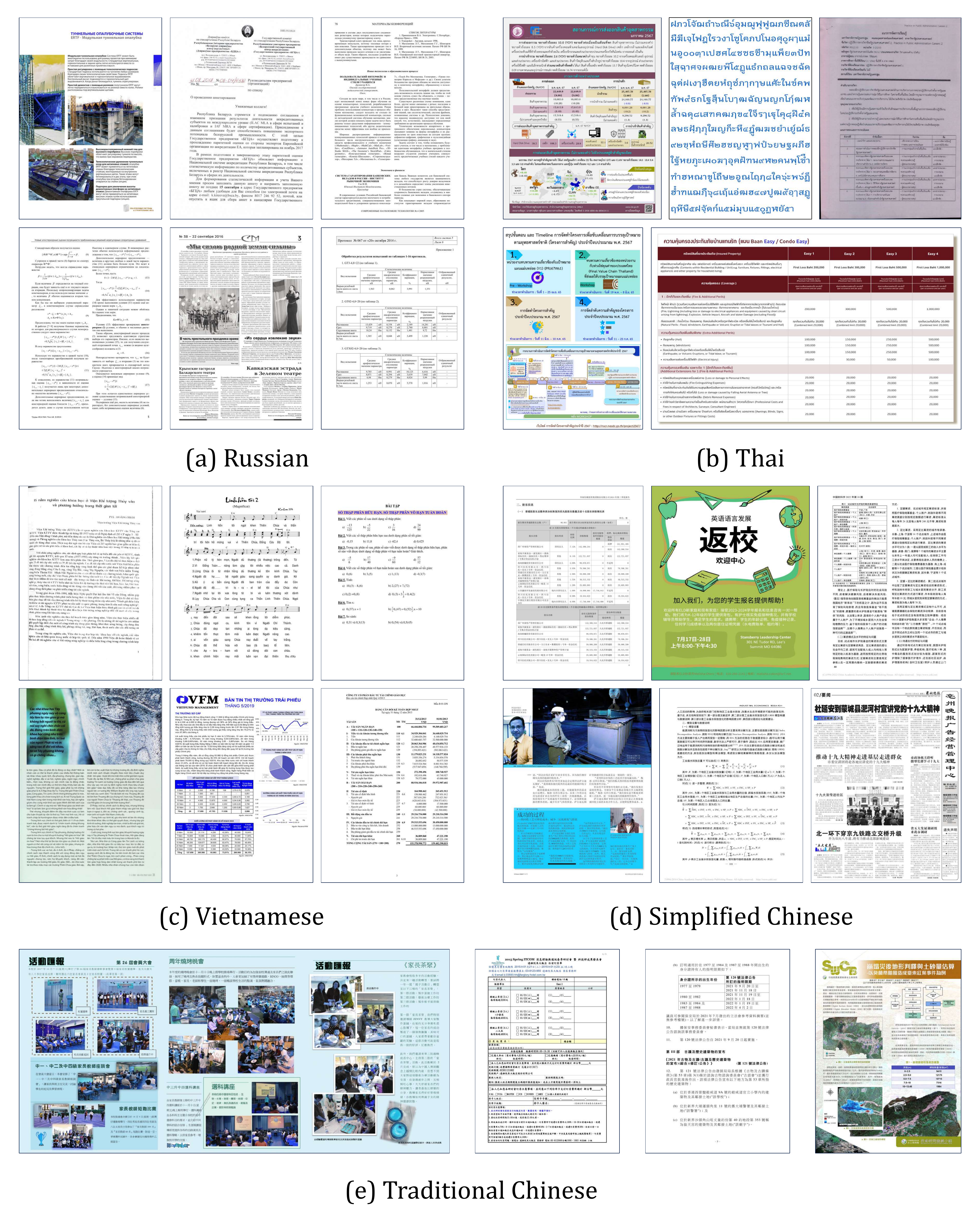}
    \caption{Visualization of Russian, Thai, Vietnamese, Simplified Chinese, and Traditional Chinese images in MonkeyDoc v2.}
    \label{fig:app-e-training-examples-group3}
\end{figure}

\end{document}